\documentclass{article}

    \PassOptionsToPackage{numbers, compress}{natbib}
 \usepackage[main, final]{neurips_2026}
\usepackage[utf8]{inputenc}
\usepackage[T1]{fontenc}
\usepackage{hyperref}
\usepackage{url}
\usepackage{booktabs}
\usepackage{amsfonts}
\usepackage{nicefrac}
\usepackage{microtype}
\usepackage{xcolor}

\newcommand\mypara[1]{\noindent\textbf{#1.}}
\usepackage{soul}           % underline with \ul{}
\usepackage{enumitem}       % enumerate
\usepackage{graphicx}
\usepackage{amsmath}
\usepackage{listings}          % provides \lstinputlisting
\usepackage{xspace}
\usepackage{subcaption}
\usepackage{bbold}      % full alphabet + digits, but changes \mathbb everywhere
\usepackage{bbm}        % use \mathbbm{1} for the indicator function
\usepackage{wrapfig} % in preamble
\usepackage[normalem]{ulem}

\usepackage{colortbl}
\usepackage{xcolor}
\newcommand{\bestinline}[1]{{\setlength{\fboxsep}{1pt}\colorbox{gray!40}{\textbf{#1}}}}
\newcommand{\secondinline}[1]{{\setlength{\fboxsep}{1pt}\colorbox{gray!15}{#1}}}
\newcommand{\best}[1]{\cellcolor{gray!40}\textbf{#1}}
\newcommand{\second}[1]{\cellcolor{gray!35}#1}

\newcommand{\meanstd}[2]{%
  \shortstack{#1\\[-1pt]{\scriptsize$\pm$ #2}}%
}

\newcommand{\ourmethod}{HorizonComposer\xspace}

\title{\ourmethod: Spatiotemporally Consistent Driving Video Editing with Enriched Traffic Semantics}

\author{%
    Mauricio Soroco\thanks{Equal contribution.} \\
    Simon Fraser University \\
    \And
    Yuqiu Liu\footnotemark[1] \\
    Simon Fraser University \\
    \And
    Zaid Tasneem \\
    NEC Labs America \\
    \And
    Francesco Pittaluga \\
    NEC Labs America \\
    \And
    Abhishek Aich \\
    NEC Labs America \\
    \And
    Wuyang Chen \\
    Simon Fraser University \\
    \texttt{wuyang@sfu.ca} \\
    \And
    Manmohan Chandraker \\
    NEC Labs America \\
    \And
    Ziyu Jiang \\
    NEC Labs America \\
    \texttt{zjiang@nec-labs.com}
}

\begin{document}

\maketitle

\begin{abstract}
  Instruction-guided video generative models offer a scalable solution for stress-testing autonomous driving systems by simulating diverse environmental conditions.
  However, 
   driving scenes operate within a dense Operational Design Domain (ODD) governed by strict traffic semantics. 
  When applied to global driving-condition edits,
  general-purpose instruction-guided video editors
  frequently corrupt these semantics, leading to structural hallucinations and ``temporal shock.''
  This work distills and protects traffic semantics during video editing at three levels.
  First, we construct a hybrid dataset by combining temporally aligned real-world videos with synthetic edits, balancing foundational scene structure with unambiguous appearance signals.
  Second, during training, we adapt advantage conditioning from offline reinforcement learning to align the model using dense rewards.
  We extract dense, pixel-level rewards from the hybrid data to serve as a localized signal and
  actively steer generation
  toward structurally reliable pixels.
  Finally, our model mitigates temporal shock by introducing thinking frames, a transition mechanism that seamlessly reconciles static image conditioning with dynamic video context.
  Extensive experiments demonstrate that our method prevents the corruption of multi-actor dynamics, establishing state-of-the-art spatiotemporal consistency,
  yielding consistent user preference (\textbf{39\%} relative improvement in content preservation and \textbf{12\%} in temporal consistency),
  and achieving a significant \textbf{5\%} gain in downstream road line segmentation and \textbf{7\%} gain in downstream multi-object detection.
\end{abstract}

\section{Introduction}

\begin{figure}[ht]
    \centering
    \includegraphics[width=\textwidth]{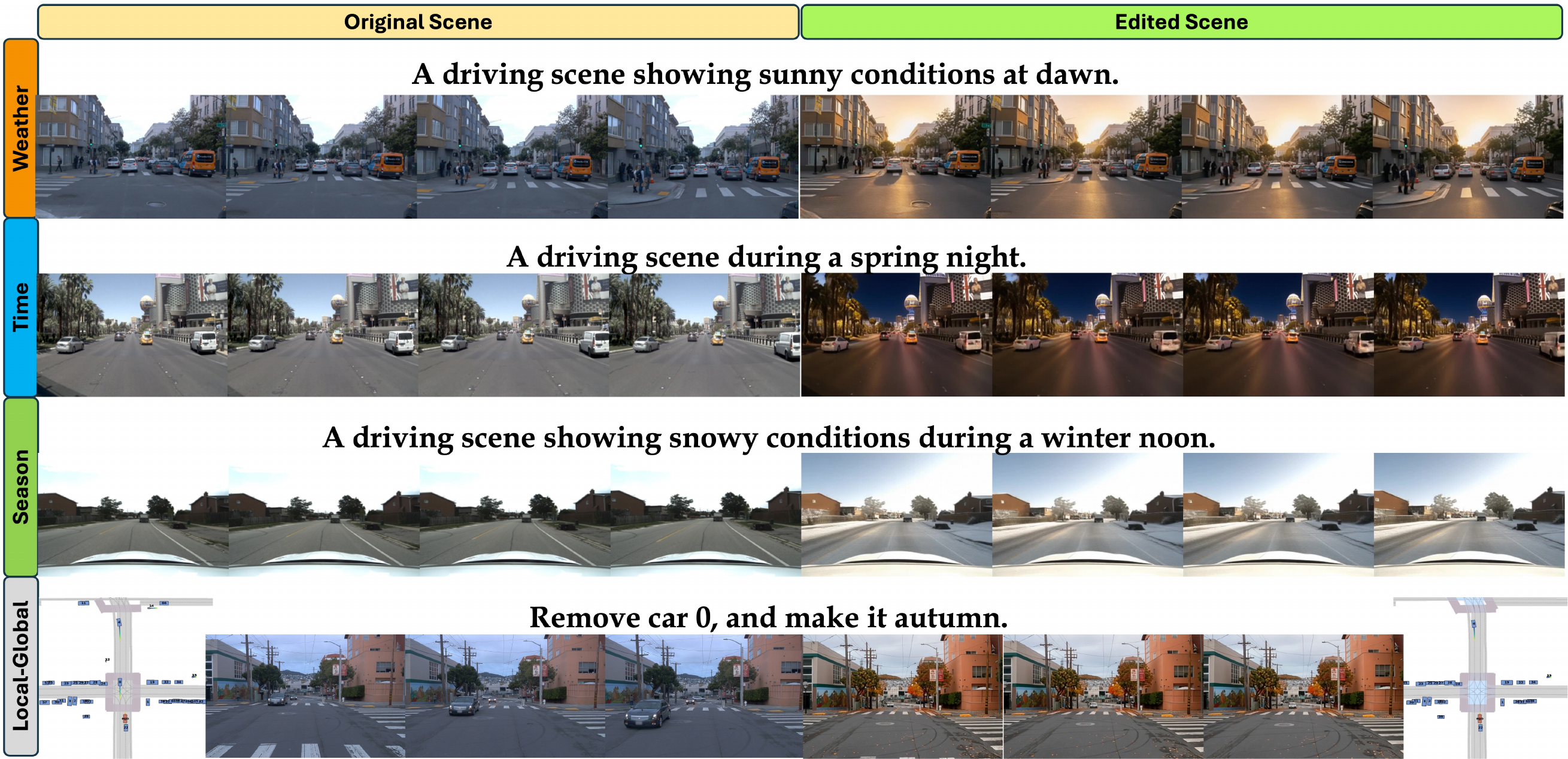}
    \caption{\ourmethod maintains complex traffic semantics during global appearance editing (rows 1-3) using a careful design across data (Sec~\ref{sec:data}), model (Sec.~\ref{sec:image-robustness}), and training (Sec.~\ref{sec:rewards}). Beyond global transformations, our fidelity in traffic semantics allows \ourmethod to seamlessly compound global and local edits for comprehensive scene simulation (row 4). 
    } 
    \label{fig:teaser}
\end{figure}

Autonomous driving systems must operate safely and reliably across a vast Operational Design Domain (ODD), encompassing diverse weather conditions, seasonal variations, and day-night cycles. 
To robustly evaluate perception and planning algorithms, we require scalable simulation environments capable of simulating diverse conditions for the exact same driving scene.
However, collecting perfectly paired, real-world data across these variations is practically impossible due to the natural evolution of traffic patterns.
Consequently, instruction-guided video generative models (VGMs) have emerged as a promising proxy~\citep{nvidia2026worldsimulationvideofoundation}, offering the ability to controllably transform the global appearance of existing driving logs (e.g., weather or time) to stress-test autonomous algorithms.

Unlike object-centric videos, driving operation design domains are defined by the dense co-existence of numerous critical factors in a single frame: multiple independent actors, strict lane geometries, and complex infrastructure. 
This spatiotemporal density forms a coupled and constrained set of \textit{traffic semantics}.
Because of this inherent complexity, editing driving scenes requires a delicate balance between \textit{Instruction Alignment} (altering global appearance) and strictly enforcing \textit{ODD Preservation} (maintaining local scene structure). 
In practice, global appearance editing almost always corrupts local traffic semantics, leading to severe spatiotemporal inconsistencies. Obtaining precise video-editing pairs to supervise driving scene editing is challenging, leaving the training of instruction-guided video editors hampered by three critical bottlenecks:

\mypara{1. Unpaired Supervision Leads to Semantic Corruption}
Controlled real-world driving video (e.g., the same route captured in different weather) is rare and suffers from temporal unpairing due to dynamic traffic flow.
Training naively on these unpaired videos misleads the model into believing that vehicle identities and locations can arbitrarily change during an appearance edit. We require a mechanism to definitively separate the desired global appearance shifts from the strict local structure that must be preserved.

\mypara{2. Misaligned Synthetic Priors}
Conversely, using baseline models to generate synthetic real-world editing pairs provides strong appearance signals but may introduce structural hallucinations or "creative additions" that violate the ODD. Because the supervision quality varies drastically across pixels and frames, a robust training paradigm cannot treat all regions equally.

\mypara{3. Conditioning-induced Temporal Shock}
Video generators benefit from image-conditioning signals that guide global appearance.
However, in dense traffic scenes, these static reference images frequently suffer from actor misalignment and are not guaranteed to be consistent with the underlying video dynamics. 
Injecting these conditions directly into the generation pipeline leads to "temporal shock": unnatural transitions, structural warping, and flickering artifacts, which are much more serious than temporal inconsistency.

These bottlenecks motivate our core question:
\begin{center}\label{sec:intro-core-question}
\fbox{
    \parbox{0.86\linewidth}{
        \textit{How to distill rich and faithful traffic semantics into driving scene video editing and thus achieve spatiotemporal consistency?}
    }
}
\label{core_questions}
\end{center}

To address this question, we propose \ourmethod, a three-pronged approach to inject and protect traffic semantics throughout driving video editing:

\mypara{Data - Pairing Hybrid Videos} We construct a hybrid dataset by pairing temporally aligned real-world driving videos with synthetic edits to introduce controlled appearance variations. While real-world pairs preserve the foundational scene structure despite traffic inconsistencies, synthetic pairs provide unambiguous appearance signals. 

\mypara{Training - Aligned Semantics via Dense Rewards}
We align our editor with the spatiotemporal density of driving scenes through offline reinforcement learning, extracting \textit{dense, pixel-level rewards} from the hybrid data as a localized supervision signal that steers generation toward structurally reliable pixel states.

\mypara{Model - Temporally Robust Transition} 
To mitigate temporal shock, we introduce \emph{thinking frames}, a mechanism that provides a temporal transition between the static conditioning signal and the dynamic video content. 
By prepending thinking frames during inference, the model can reconcile undesired artifacts to smoothly incorporate the intended edit without corrupting multi-agent dynamics.

We summarize our concrete contributions below:
\begin{enumerate}[leftmargin=*]
\setlength\itemsep{0em}

\item \textbf{Comprehensive Datasets for Driving Video Editing.}
We construct a comprehensive editing dataset for instruction-guided driving video edits, featuring paired real and synthetic videos with corresponding editing masks and prompts.
In total, we collect 86614 videos covering 172 categories of driving-scene semantics.

\item \textbf{Superior Driving Video Editing Performance.}
\ourmethod outperforms recent video editing methods for safety-critical structural fidelity on unseen driving scenarios, achieving up to 20\% improvements in content preservation while maintaining competitive instruction alignment and photorealism.

\item \textbf{Consistent User Preference.}
\ourmethod consistently generated driving video edits preferred by real users (via voting), with a relative out-performance over the second-best method by 21 \% overall.

\item \textbf{Benefits to Downstream Driving Tasks.}
Including driving scene videos edited by \ourmethod further boosts downstream road line segmentation performance by 5\% and boosts downstream multi-object detection by up to 7.14\%.

\end{enumerate}

\section{Related Work}
\label{sec:related}

\mypara{Instruction-guided video editing} 
General-purpose video editors built on large generative backbones~\citep{wan2025wanopenadvancedlargescale, hacohen2024ltxvideorealtimevideolatent, kong2025hunyuanvideosystematicframeworklarge} and unified MLLM-diffusion frameworks~\citep{wei2025univideo, mou2025instructxunifiedvisualediting, jiang2025vaceallinonevideocreation, ju2026editverseunifyingimagevideo, chen2026vinounifiedvisualgenerator} have made rapid progress, but they offer no native guarantee on the strict spatiotemporal traffic semantics that driving requires. The work most relevant to us is ChronoEdit~\citep{wu2025chronoedittemporalreasoningimage}, whose intermediate ``reasoning'' frames inspire our \emph{thinking frames}; however, we adapt the idea to anchor multi-agent dynamics rather than single-image physics.

\mypara{Driving scene synthesis and editing} 
A growing body of work generates driving footage from layouts or text~\citep{swerdlow2024BEVGen,yang2023bevcontrol,wang2023drivedreamer,gao2023magicdrive,wen2024panacea,gao2024magicdrivedit,lu2024infinicube,chen2025autoscape,alhaija2025cosmos,nvidia2026worldsimulationvideofoundation}, while a parallel line edits real logs through 3D/4D priors, particle fields, or occupancy~\citep{wang2026horizonforge,soroco2026horizonweaver,ljungbergh2025r3d2,qian2025weathereditcontrollableweatherediting,mo2025dreamlandcontrollableworldcreation,li2025unisceneunifiedoccupancycentricdriving,schneider2025neuralatlasgraphsdynamic}. These approaches typically rely on explicit reconstruction or external simulators and still struggle to preserve original vehicle identities under appearance edits. We instead operate directly in the video domain and enforce ODD preservation through dense pixel-level rewards, avoiding any 3D/4D reconstruction overhead.

\mypara{Reinforcement learning for diffusion} 
On-policy methods~\citep{liu2025flowgrpotrainingflowmatching, zhao2026imageeditr1boostingmultiagentimage, black2024trainingdiffusionmodelsreinforcement, li2024instructrl4pixtrainingdiffusionimage} are prohibitive at video scale in terms of computing cost, while offline weighted regression~\citep{peng2019advantageweightedregressionsimplescalable} collapses to filtered imitation~\citep{intelligence2025pi06vlalearnsexperience}; recent inference-time guidance methods~\citep{jin2025inferencetimealignmentcontroldiffusion,frans2025diffusion, cheng2026diffusionblendinferencetimemultipreference} sidestep this but rely on scalar, sample-level signals that miss intra-frame structural variation. We extend this line with dense, pixel-aligned reward indicators that extract localized policy improvements from partially reliable hybrid data.

\noindent For detailed survey, see Appendix~\ref{app:detailed-related-works} for extended discussion.

\section{\ourmethod}

\subsection{Overview} \label{sec:overview}

\ourmethod strictly distills and protects traffic semantics during video editing through interventions at the data, training, and model levels. Sec.~\ref{sec:data} outlines our data pipeline, constructing a hybrid dataset of real-world and synthetic video pairs to balance structure and appearance. Sec.~\ref{sec:rewards} details our training paradigm, where we extract spatiotemporal rewards from our hybrid data for reinforcement learning.
Sec.~\ref{sec:image-robustness} introduces temporal conditioning mechanisms that buffer against misaligned reference frames, ensuring seamless appearance transitions and strict actor fidelity.

\mypara{Preliminary}
We instantiate our framework on top of Cosmos-Transfer2.5~\cite{nvidia2026worldsimulationvideofoundation}, a diffusion transformer (DiT) that performs latent-space video denoising under multiple structured control branches. Given a noisy latent $\mathbf{z}_\tau$ at diffusion timestep $\tau$, the base model $\epsilon_\theta$ predicts a denoising velocity conditioned on a text prompt $\ell$,
a reference image $\mathbf{c}_{\text{img}}$, and a stack of spatial control signals $\mathbf{c}_{\text{ctrl}} = \{\mathbf{c}^{(1)}, \dots, \mathbf{c}^{(K)}\}$ (e.g., depth, visual) injected through dedicated ControlNet-style branches. Of these, we retain two: a depth branch that anchors the geometric layout of the source clip, and a visual branch that takes the source clip itself as input to carry its appearance prior. We freeze the depth branch to strictly enforce structural fidelity, and finetune the visual branch to specialize it for driving-scene editing.

For a source driving clip $\mathbf{x}^s = \{\mathbf{x}^s_0, \dots, \mathbf{x}^s_{T-1}\}$ and a target editing instruction $\ell^t$, our goal is to produce an edited clip $\mathbf{x}^t$ that (i) aligns globally with $\ell^t$ and (ii) preserves the local traffic semantics of $\mathbf{x}^s$.
We denote our editing model as $\mu_\theta(\mathbf{x}^t \mid \mathbf{o})$, where the observation $\mathbf{o} = (\mathbf{x}^s, \mathbf{c}^{(0)}_{\text{img}}, \mathbf{c}_{\text{depth}}, \ell^t)$
bundles the source clip, a reference image generated by an image editor~\cite{google_nanobanana}, source depth sequence, and the target instruction.

\begin{figure}[ht]
    \centering
    \includegraphics[width=\textwidth]{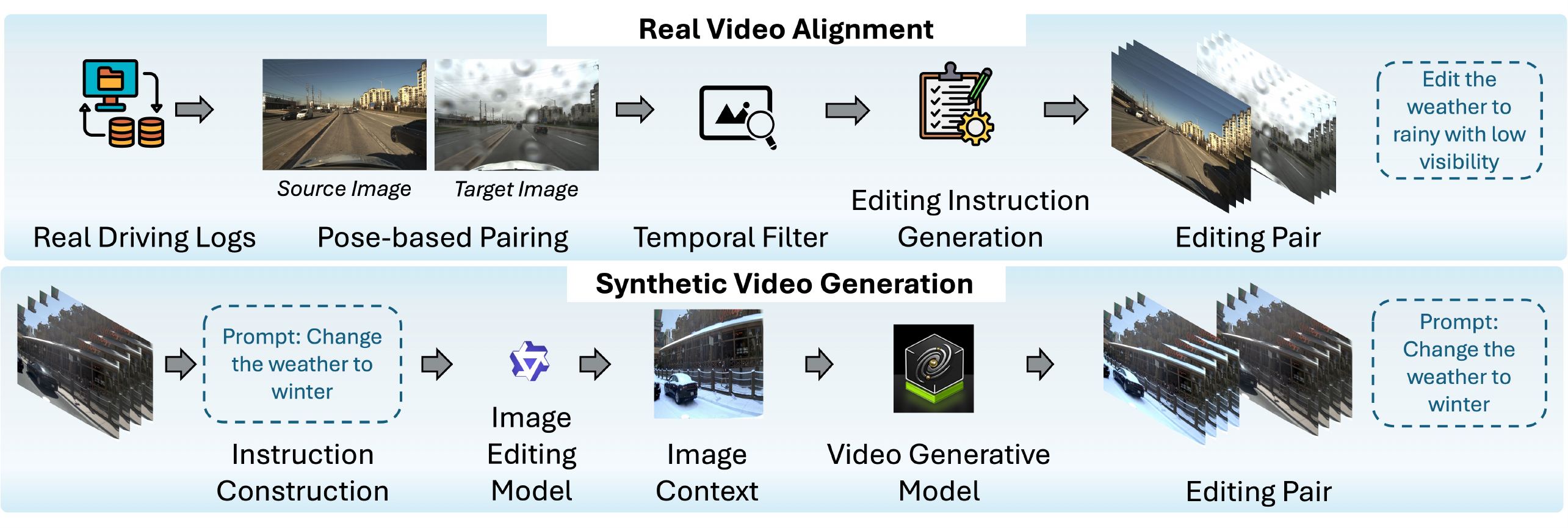}
    \caption{\textbf{Dataset Construction and Pseudo-Labeling Pipeline.} \textbf{(Top)} \textit{Realistic Dataset Development} : Real-world data (Sec.~\ref{sec:data-real-world}) are matched via pose alignment, temporally filtered, and processed to generate editing instructions, forming real-world video pairs. \textbf{(Bottom)} \textit{Pseudo Dataset Development} (Sec.~\ref{sec:data-pseudo-gt}): We synthesize driving scenes by editing the initial frame with an image prior, followed by a video generative model to propagate the edits and produce synthetic video pairs. 
    }
    \label{fig:dataset-construction-pipeline}
\end{figure}

\subsection{Distilling Semantics via Hybrid Video Pairing}\label{sec:data}

\subsubsection{Real-World Data Harvesting through Structural Alignment}\label{sec:data-real-world}

We begin by introducing our real-world data pipeline, which produces high-quality, temporally consistent, and well-aligned video pairs from large-scale driving datasets.
We build on Boreas~\cite{burnett_ijrr23}, a dataset of multi-season driving logs collected over repeated routes.
As illustrated in Figure~\ref{fig:dataset-construction-pipeline}, our pipeline transforms raw video logs into temporally aligned pairs spanning diverse environmental conditions through three stages:

\begin{enumerate}[leftmargin=*, topsep=0pt, partopsep=0pt, itemsep=2pt, parsep=0pt]
\setlength\itemsep{0em}

\item \emph{Pose-based Pairing.}
We first pair frames from different traversals of the same route using camera pose. Specifically, we designate one log as the source video and, for each of its frames, retrieve the nearest frame from another traversal of the same route captured on a different day.
This yields two viewpoint-matched sequences recorded under different weather, seasons, or times of day.

\item \emph{Temporal Filtering.}
However, viewpoint matching alone does not guarantee temporal consistency, since the two traversals exhibit independent traffic dynamics.
For example, when approaching an intersection, the ego-vehicle in one traversal may stop at a red light while the other proceeds through.
Naively pairing these sequences forces one clip to dwell on a near-duplicate frame until the stationary ego-vehicle resumes motion, producing pairs in which one side is effectively static.
To address this, we apply a temporal filter that detects and removes both freeze-frame segments and segments with large timestamp gaps, then splits the remaining footage into fixed-length clips with consistent ego-motion and uniform duration.

\item \emph{Instruction Generation.}
Finally, we use a vision-language model to extract scene-level attributes from each clip (e.g., weather, lighting, road layout), and convert the difference between source and target attributes into a natural-language editing instruction.
\end{enumerate}

The resulting corpus consists of real-world video pairs in which the underlying scene structure is preserved while global appearance varies, providing strong supervision for structure-faithful driving-scene editing. More details can be found in App.~\ref{app:real-world-data}

\subsubsection{Synthetic Appearance Generation}\label{sec:data-pseudo-gt}

Collecting real-world videos along repeated routes is expensive, and even aligned traversals cannot preserve identical traffic dynamics. To complete this missing supervision, we turn to synthetic appearance-editing pairs. However, we find that available state-of-the-art video editing models often yield low-quality or unstable results on driving scenes. We therefore construct synthetic pairs using a two-stage pipeline inspired by CCEdit~\cite{feng2024ccedit} (first edit a reference frame, then propagate the edit to the full video) but powered by stronger components: Qwen Image Editing~\cite{wu2025qwenimagetechnicalreport} for first-frame editing, and Cosmos-Transfer2.5~\cite{nvidia2026worldsimulationvideofoundation} for video propagation.

For propagation, we feed both depth and visual signals into Cosmos-Transfer2.5 with branch weights of $0.9$ and $0.1$, respectively. The depth branch provides a strong geometric prior, while the visual branch contributes appearance cues from the source clip. We deliberately keep the visual weight small: a larger weight causes the model to over-anchor to the source appearance, producing videos that ignore the target edit entirely. We use Qwen here rather than Nano Banana for cost efficiency given the scale of training-data construction; at inference time and elsewhere in our pipeline, we use Nano Banana~\cite{google_nanobanana} as the reference image editor (see Sec.~\ref{sec:image-robustness} and Sec.~\ref{sec:settings}). Full implementation details and dataset statistics are provided in App.~\ref{app:synthetic_generation} and App.~\ref{app:dataset_statistics}, respectively. 

It is worth noting that this synthetic generation pipeline closely mirrors the architecture of \ourmethod itself. Initializing from a framework that already performs a closely related task lets us inherit its pretrained knowledge and yields stronger generalization than training from scratch.

\subsection{Semantics Alignment with Fine-grained Reward and Offline-RL Fine-tuning}\label{sec:rewards}

\begin{wrapfigure}{r}{0.6\textwidth}
    \centering
    \includegraphics[width=\linewidth]{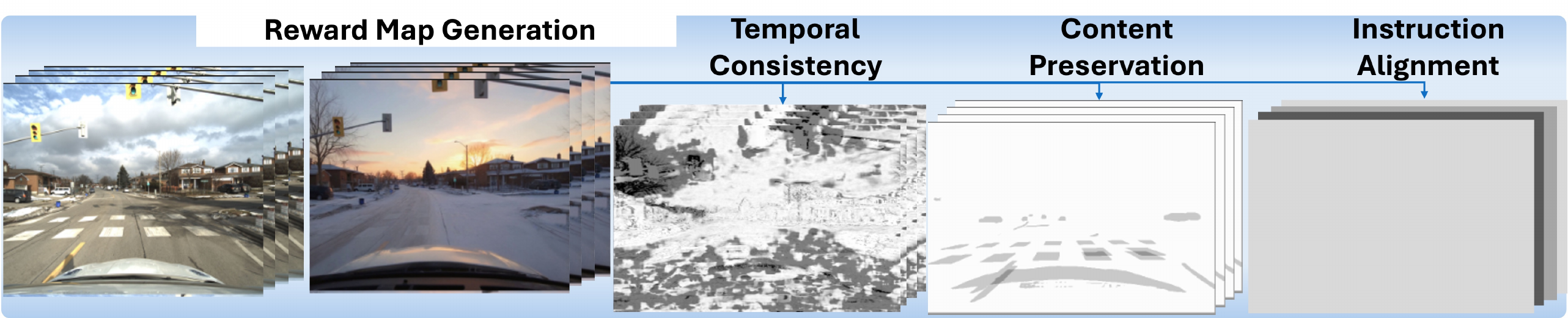}
    \caption{\textbf{Fine-grained Reward Map Generation} (Sec.~\ref{sec:data-rewards}): For each video pair, we combine frame-level video-editing metrics, pixel-level scene-change masks and driving-specific semantic checks to construct dense reward maps, which guide training toward reliable appearance edits while protecting safety-critical traffic semantics. Technical details are in App.~\ref{app:rewards}.
    }
    \label{fig:reward-construction-pipeline}
\end{wrapfigure}
Our hybrid paired dataset $\mathcal{D}$ (Section ~\ref{sec:data}) provides rich traffic semantics supervision (App.~\ref{app:training-inference-details}), but neither source is perfect.
Real-world pairs preserve scene structure yet exhibit unaligned traffic between traversals,
while synthetic pairs deliver clean appearance signals yet may hallucinate or distort safety-critical semantics like lane markings.
Since these artifacts are spatially sparse yet appear in nearly every video, conventional sample-level filtering (i.e., discarding "bad" videos) would decimate the dataset or admits pervasive local corruption.
To exploit reliable signals while suppressing artifacts, the supervision must be grounded at the pixel level.
Section~\ref{sec:data-rewards} describes how we construct dense reward maps that combine video-editing metrics with driving-specific semantic checks to score each pixel's reliability. Section~\ref{sec:offline_RL} formulates training as offline reinforcement learning to maximize appearance changes in reliable regions while strictly protecting vulnerable traffic semantics.

\subsubsection{Constructing Fine-grained Traffic-Semantic Rewards} \label{sec:data-rewards}

We construct a dense reward map $\mathbf{r} \in [-1,1]^{3 \times H \times W \times T}$ that assigns per-pixel reliability scores to each training pair along three complementary axes (Fig.~\ref{fig:reward-construction-pipeline}): \textbf{Content Preservation}, \textbf{Temporal Consistency}, and \textbf{Instruction Alignment}. Aggregated together, these axes yield the improvement indicator $I$ used in Sec.~\ref{sec:offline_RL}.

\textbf{Content Preservation} anchors traffic semantics by assigning low rewards to regions where source content is hallucinated or deleted. Rather than relying on simple pixel diffs, we utilize a SAM-based scene change detector~\citep{kim2025generalizablescenechangedetection} to identify structural deviations. Also, we explicitly segment safety-critical elements like traffic lines, crosswalks, and vehicles using SAM3~\citep{carion2025sam3segmentconcepts}, penalizing any geometric corruption. Furthermore, we employ a Vision-Language Model~\citep{chen2024internvl} to verify that original vehicle appearance is preserved.

\textbf{Temporal Consistency} measures structural fidelity across frames to penalize flickering and motion artifacts. We compute this using FF-$\alpha$ and FF-$\beta$ \citep{chen2025editboardcomprehensiveevaluationbenchmark}, which track intensity-based consistency via backward-warped optical flow and motion orientation, respectively, relative to the source video.

\textbf{Instruction Alignment} evaluates how faithfully the generated frames reflect the target editing instruction $\ell^t$. We compute a success rate using CLIP cosine similarity, comparing each frame against both the source and target prompts. Because this is a frame-level scalar metric, its value is broadcast spatially across all pixels for the given frame.

We defer per-metric definitions, prompts, and implementation details to App.~\ref{app:rewards}.

\subsubsection{Offline Reinforcement Learning}
\label{sec:offline_RL}

While neither real-world nor synthetic pairs are perfect, we treat them jointly as off-policy rollouts of an imperfect behavior policy $\mu_{\text{ref}}$ defined by our hybrid dataset $\mathcal{D}$, and employ an offline RL approach to learn an improved policy $\mu_\theta$ from this data. We define $\mathbf{o}$ as the conditioning inputs (i.e., the source video and text prompt) and the action $\mathbf{a} \equiv \mathbf{x}^t$ as the generated edited video clip.

\mypara{CFGRL offline RL} We build on CFGRL~\cite{frans2025diffusion}, an offline RL framework that performs policy improvement via advantage conditioning. Under the standard regularized RL objective, the optimal improved policy admits $\hat{\mu}(\mathbf{a}\mid\mathbf{o}) \propto \mu_{\text{ref}}(\mathbf{a}\mid\mathbf{o}) \exp(A^{\mu_{\text{ref}}}(\mathbf{o},\mathbf{a})/\beta)$,
where $A^{\mu_{\text{ref}}}$ is the advantage of action $\mathbf{a}$ under $\mu_{\text{ref}}$, and the $\beta$ hyperparameter is the KL-regularization weight. Following~\cite{frans2025diffusion, intelligence2025pi06vlalearnsexperience}, a related product-policy formulation can be derived using $I = \mathbb{1}[A^{\mu_{\text{ref}}} > \epsilon]$ as
\begin{equation}
\hat{\mu}(\mathbf{a}\mid\mathbf{o}) \propto \mu_{\text{ref}}(\mathbf{a}\mid\mathbf{o}) \left( \frac{\mu_{\text{ref}}(\mathbf{a}\mid I, \mathbf{o})}{\mu_{\text{ref}}(\mathbf{a}\mid\mathbf{o})} \right)^{\beta},
\label{eq:cfgrl-policy}
\end{equation}
which avoids the need to explicitly model the advantage.
The improved policy $\hat{\mu}$ is approximated by training $\mu_\theta$ to represent both the unconditional and indicator-conditional distributions. Following ~\cite{frans2025diffusion}, we randomly drop $I$ during training to learn the unconditional prior. 
\begin{equation}
\min_\theta \; \mathbb{E}_{\mathcal{D}}\Big[ -\log \mu_\theta(\mathbf{x}^t \mid \mathbf{o}) - \alpha \log \mu_\theta(\mathbf{x}^t \mid I, \mathbf{o}) \Big].
\label{eq:cfgrl-loss}
\end{equation}
where $\alpha$ is a trade-off hyperparameter following~\cite{intelligence2025pi06vlalearnsexperience}. 
At inference, sampling with $I = 1$ favors outputs with higher expected rewards from $\hat{\mu}$ than the behavior policy~\cite{frans2025diffusion} 
(additional details in App.~\ref{app:offline_RL_expanded}).

\mypara{Offline RL with dense signal} Standard CFGRL applies a single scalar indicator $I \in \{0,1\}$ to gate an entire action. This is appropriate for low-dimensional action spaces with sparse, episodic rewards, but driving video editing inverts both assumptions:
the action $\mathbf{x}^t$
is a high-dimensional pixel volume.
We therefore extend the indicator to a dense, pixel-aligned mask $I \in \{0,1\}^{H \times W \times T}$ derived from the reward map of Sec.~\ref{sec:rewards} via $I_t(\mathbf{p}) = \mathbb{1}[\mathbf{r_t}(\mathbf{p}) > \epsilon]$.
Each pixel is independently labeled positive or negative, allowing $\mu_\theta$ to extract policy improvement signal from reliable regions of an otherwise noisy training example while learning to avoid the failure modes encoded in negative pixels.
Crucially, this dense formulation requires no architectural change to Eq.~\ref{eq:cfgrl-loss}, with the indicator simply broadcast to pixel resolution, making the approach scalable to higher resolutions, longer videos, or additional reward channels without modifying the underlying training objective.

\mypara{Implementation} 
The dense indicator $I$ is encoded as an additional control branch and injected through the visual control branch alongside the depth and reference-image conditions. Following~\cite{intelligence2025pi06vlalearnsexperience}, the log-likelihoods in Eq.~\ref{eq:cfgrl-loss} are realized as the standard flow-matching loss of the base model.
During training, we drop $I$ with probability 0.3 to retain the unconditional distribution. 
Having learned to associate $I=1$ with high-quality structural preservation during training, the model is conditioned at inference by setting $I = \mathbf{1}$ globally, to treat the entire spatial volume as a high-quality target.

\subsection{Thinking Frames for Smoothing Temporal Shock}\label{sec:image-robustness} 

While the reference image $\mathbf{c}_{\text{img}}$ is powerful for guiding global appearance transitions (e.g., shifting weather or time of day), those generated by general-purpose state-of-the-art editing methods such as Nano Banana~\cite{google_nanobanana} may introduce \emph{actor misalignment and structural hallucinations}, due to the dense co-existence of elements in driving scenes and less awareness of the traffic semantics.
Naively instructing the driving video generation using these imperfect reference images often causes a "\emph{temporal shock}," as the model struggles to immediately reconcile the static reference artifacts with the strict spatiotemporal dynamics of the traffic.

To mitigate this, we introduce \textbf{Temporal Thinking Frames}.
Specifically,
\begin{enumerate}[leftmargin=*, topsep=0pt, partopsep=0pt, itemsep=2pt, parsep=0pt]
\setlength\itemsep{0em}
\item
Rather than forcing an abrupt transition from the reference image to the target video, we prepend a sequence of $N=2$ thinking frames to the temporal dimension.

\item
To align this temporal buffer with latent traffic semantics, we simultaneously prepend a temporally reflected sequence of the driving controls (e.g., depth or structural layouts).
Concretely, if the initial control frames are denoted as $\mathbf{c}_{0:N-1}$, the prepended control segment is the temporal reversal $[\mathbf{c}_{N-1}, \dots, \mathbf{c}_0]$.
This reflected control prefix creates a smooth, temporally coherent "warm-up" period at the onset of generation without introducing externally synthesized, unverified structural data.

\item
During these thinking frames, the model is forced to resolve the discrepancies between the image condition and the video condition. It gradually smooths out the actor misalignments present in the reference image, merging the desired global appearance into the rigid geometric constraints of the driving scene.

\item
After the denoising process is complete, this prepended sequence is cropped.
\end{enumerate}

 Consequently, the final video retains its original length and strict semantic layout, having fully benefited from the stabilized appearance conditioning achieved during the internal thinking steps.

\section{Experiments}

\subsection{Settings}
\label{sec:settings}

We structure our experiments to answer our core question (Sec. \ref{sec:intro-core-question}): \textit{How to distill rich traffic semantics into driving scene video editing and thus achieve spatiotemporal consistency?}
We assess editing performance based on the preservation of traffic semantics across four perspectives:
\begin{enumerate}[leftmargin=*, topsep=0pt, partopsep=0pt, itemsep=2pt, parsep=0pt]
\setlength\itemsep{0em}
\item \textbf{Temporal Consistency.}
Following the evaluation protocol from~\citep{chen2025editboardcomprehensiveevaluationbenchmark}, we evaluate FF-$\alpha\downarrow$ and FF-$\beta\downarrow$, which quantify structural fidelity and motion alignment via optical flow.

\item \textbf{Operational Design Domain (ODD) Preservation.}
We measure the cross-frame stability of foreground actors and background scenes using Subj.$\uparrow$ (DINO \citep{caron2021emergingpropertiesselfsupervisedvision} features) and Bg.$\uparrow$ (CLIP~\citep{radford2021learningtransferablevisualmodels} features). We measure road line IoU (Lane $\uparrow$) and car appearance preservation (Car $\uparrow$).

\item \textbf{Instruction Alignment.}
We report Succ.$\uparrow$, the frame-level success rate of preferring the target over the source prompt, alongside CLIP$\uparrow$ for global text-video similarity and the multimodal large language model instruction alignment score following~\citep{ku2024viescoreexplainablemetricsconditional} (GPT$\uparrow$).

\item \textbf{Photorealism.}
We employ Aes.$\uparrow$ (LAION ~\citep{schuhmann2022laion} aesthetic score) and Img.$\uparrow$ (MUSIQ~\citep{ke2021musiq} imaging quality) to evaluate visual appeal and the absence of low-level visual distortions.
\end{enumerate}

\begin{figure*}[h]
    \centering
    \hfill
    
    \begin{subfigure}[b]{0.32\textwidth}
        \centering
        \includegraphics[clip, trim=0cm 0cm 0cm 0cm, width=\textwidth]{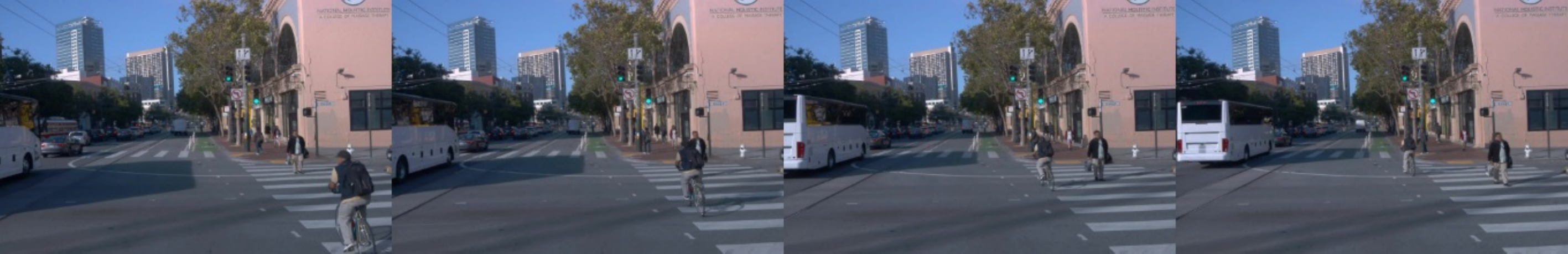}\par
        \includegraphics[clip, trim=0cm 0cm 0cm 0cm, width=\textwidth]{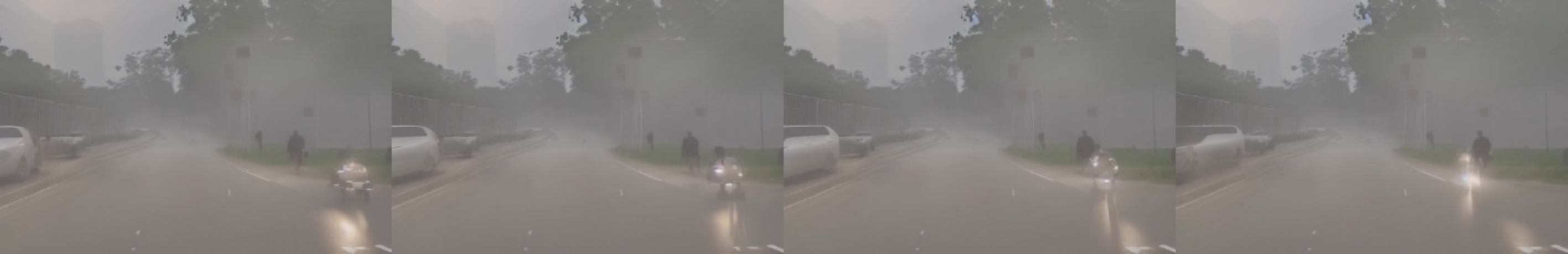}\par
        \includegraphics[clip, trim=0cm 0cm 0cm 0cm, width=\textwidth]{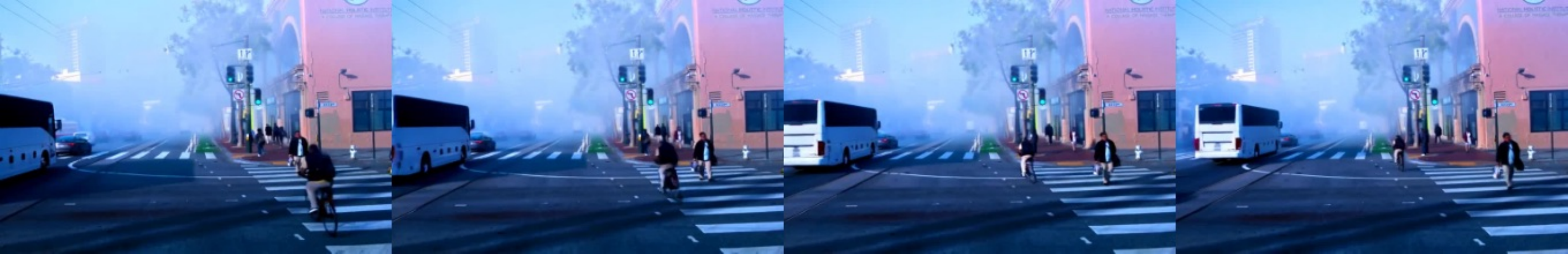}\par
        \includegraphics[clip, trim=0cm 0cm 0cm 0cm, width=\textwidth]{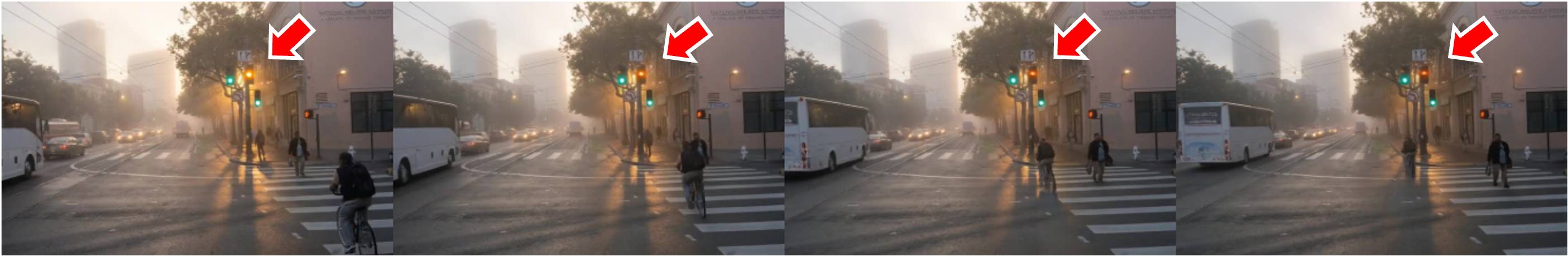}\par
        \includegraphics[clip, trim=0cm 0cm 0cm 0cm, width=\textwidth]{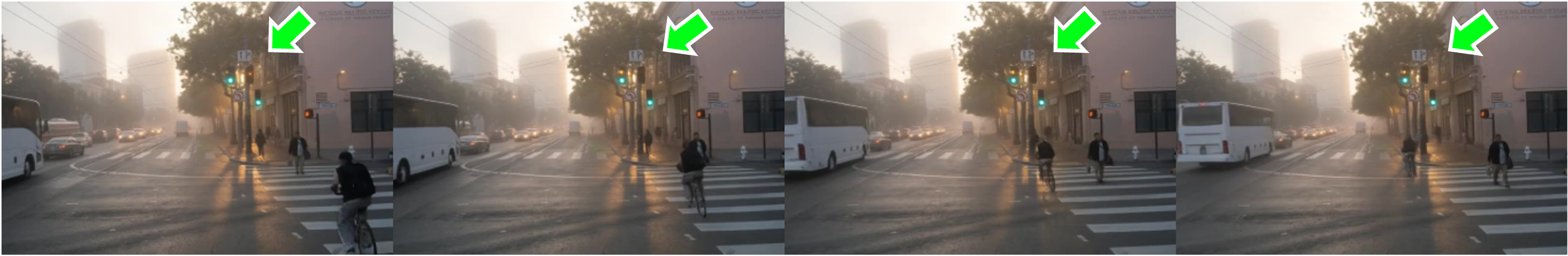}
        \caption{Sunny during a autumn daytime.}
    \end{subfigure}
    \hfill
    \begin{subfigure}[b]{0.32\textwidth}
        \centering
        \includegraphics[clip, trim=0cm 0cm 0cm 0cm, width=\textwidth]{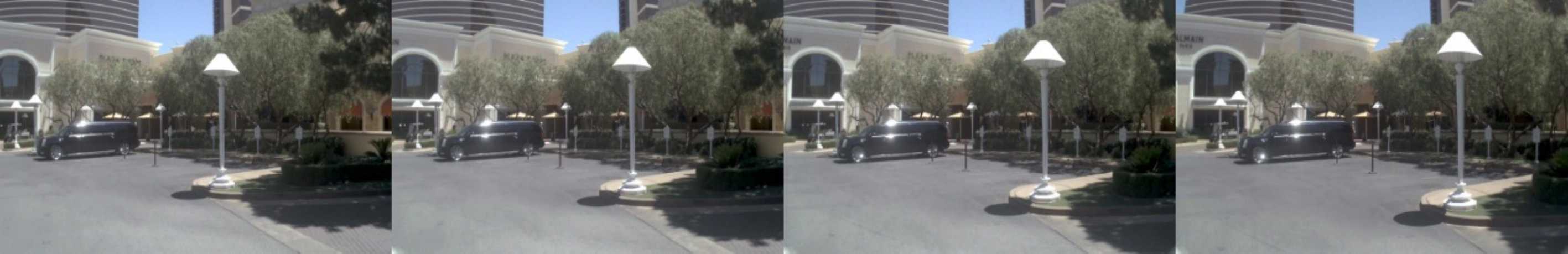}\par
        \includegraphics[clip, trim=0cm 0cm 0cm 0cm, width=\textwidth]{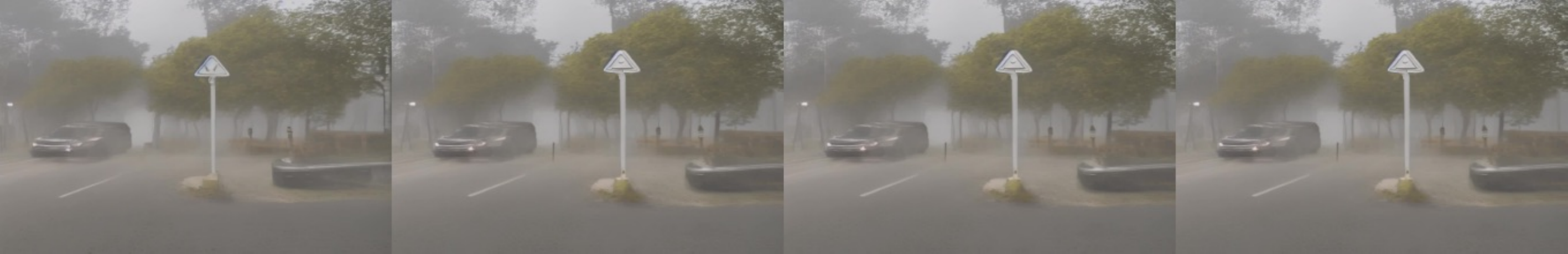}\par
        \includegraphics[clip, trim=0cm 0cm 0cm 0cm, width=\textwidth]{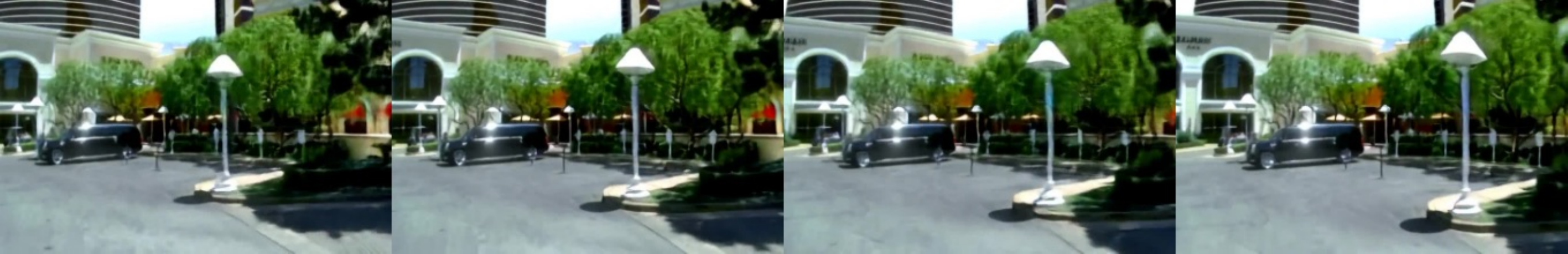}\par
        \includegraphics[clip, trim=0cm 0cm 0cm 0cm, width=\textwidth]{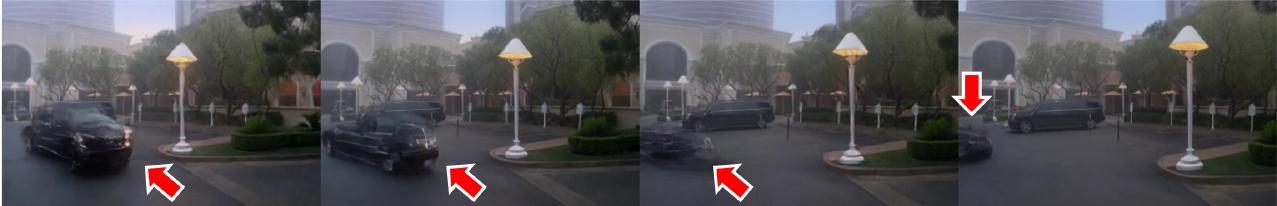}\par
        \includegraphics[clip, trim=0cm 0cm 0cm 0cm, width=\textwidth]{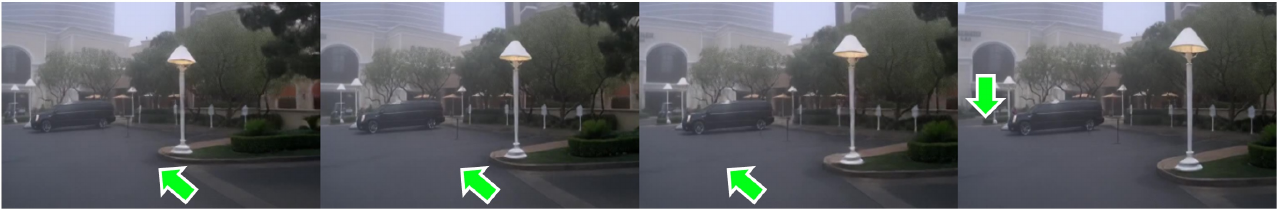}
        \caption{Foggy during a spring dawn.}
    \end{subfigure}
    \hfill
    \begin{subfigure}[b]{0.32\textwidth}
        \centering
        \includegraphics[clip, trim=0cm 0cm 0cm 0cm, width=\textwidth]{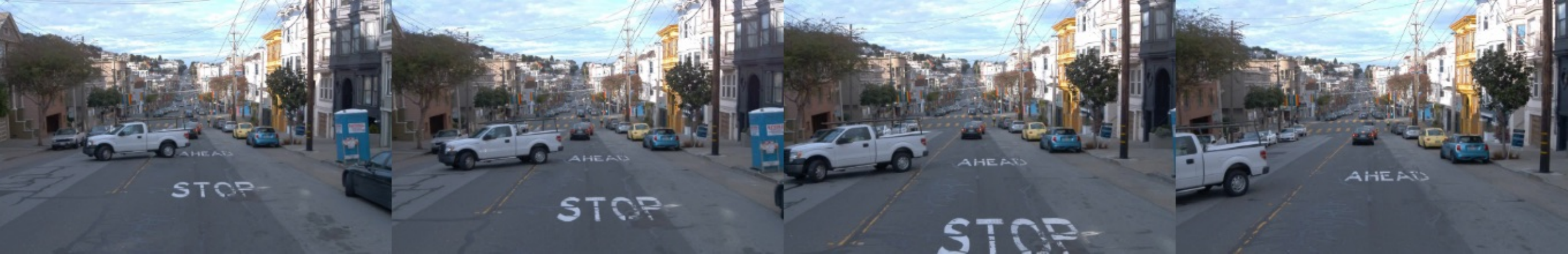}\par
        \includegraphics[clip, trim=0cm 0cm 0cm 0cm, width=\textwidth]{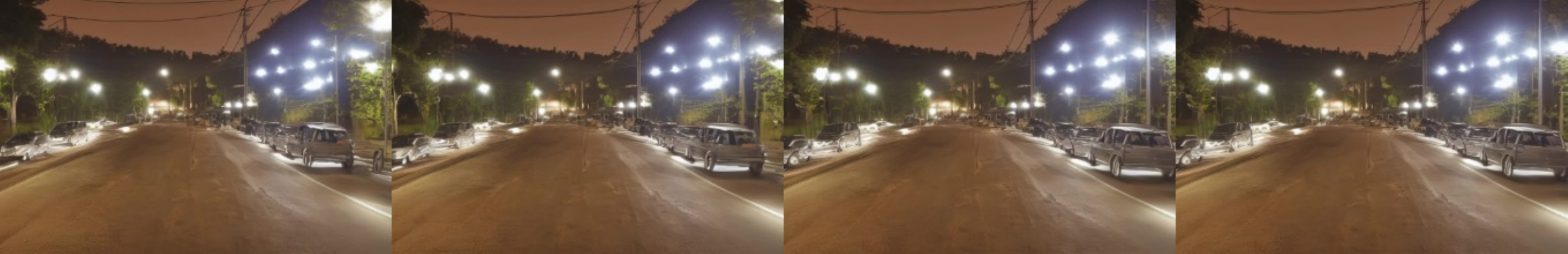}\par
        \includegraphics[clip, trim=0cm 0cm 0cm 0cm, width=\textwidth]{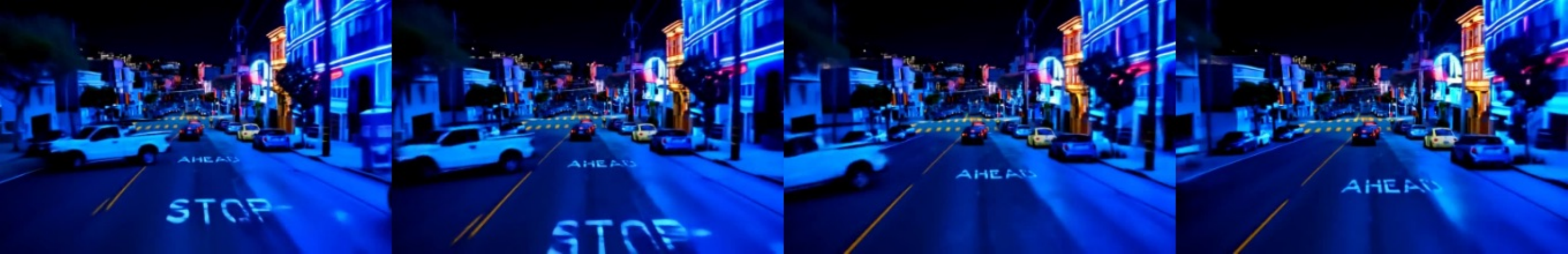}\par
        \includegraphics[clip, trim=0cm 0cm 0cm 0cm, width=\textwidth]{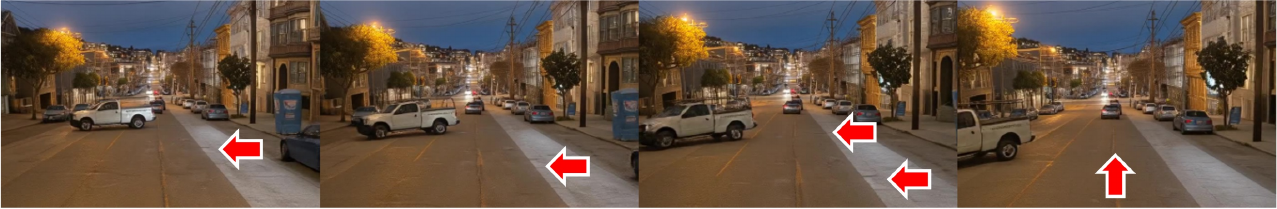}\par
        \includegraphics[clip, trim=0cm 0cm 0cm 0cm, width=\textwidth]{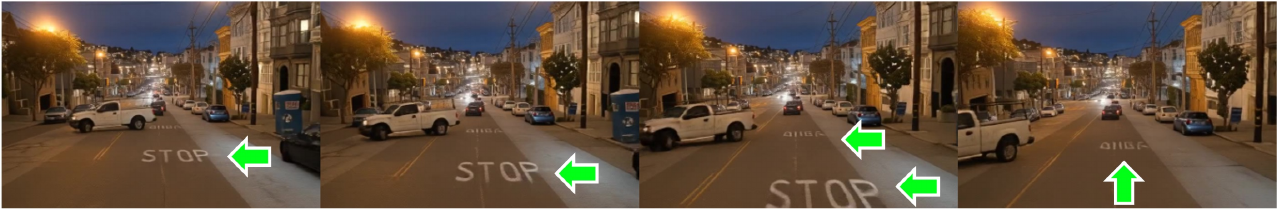}
        \caption{Night.}
    \end{subfigure}
    \hfill
    
    \caption{\textbf{Qualitative Comparison of Driving Scene Editing.} We highlight key differences in semantic preservation and visual quality across methods. Target conditions are shown in text. Row order from top to bottom: input video, CCEdit, UniVideo, SyntheticPip, Ours. \textbf{(a)} SyntheticPip hallucinates a yellow traffic light from the initial image condition. \textbf{(b)} SyntheticPip introduces a new vehicle that suddenly disappears, breaking temporal consistency. 
    \textbf{(c)} Only our method and UniVideo accurately retain the ``STOP AHEAD'' road marking. Overall, CCEdit regularly suffers from hazy artifacts and UniVideo distorts visual realism, whereas our approach maintains high-fidelity instruction alignment and strict content preservation.}
    \label{fig:exp-main-qualitative-results}
\end{figure*}

\begin{table*}[h]
\caption{\textbf{Relative Improvement (\%) over SyntheticPip.}
We report percentage improvements computed from mean scores relative
to SyntheticPip. For lower-is-better metrics, improvement is
$(b-m)/b \times 100$; for higher-is-better metrics, it is
$(m-b)/b \times 100$, where $b$ and $m$ denote baseline and model
means. Positive values indicate improvement.
We highlight \bestinline{best} and \secondinline{runner-up}
positive improvements on In-Domain (ID) and Out-of-Distribution
(OOD) scenarios. Raw scores with standard deviations are provided in Tab.~\ref{tab:main-results-raw} in the appendix.}
\centering
\small
\setlength{\tabcolsep}{3.2pt}
\renewcommand{\arraystretch}{0.9}
\begin{tabular}{l cc ccc cccc cc}
\toprule
& \multicolumn{2}{c}{\textbf{Temporal}}
& \multicolumn{3}{c}{\textbf{Edit Align.}}
& \multicolumn{4}{c}{\textbf{ODD Preservation}}
& \multicolumn{2}{c}{\textbf{Quality}} \\
\cmidrule(lr){2-3}
\cmidrule(lr){4-6}
\cmidrule(lr){7-10}
\cmidrule(lr){11-12}
\textbf{Model}
& \textbf{FF-$\alpha\downarrow$}
& \textbf{FF-$\beta\downarrow$}
& \textbf{Succ.$\uparrow$}
& \textbf{CLIP$\uparrow$}
& \textbf{GPT$\uparrow$}
& \textbf{Subj.$\uparrow$}
& \textbf{Bg.$\uparrow$}
& \textbf{Lane$\uparrow$}
& \textbf{Car$\uparrow$}
& \textbf{Aes.$\uparrow$}
& \textbf{Img.$\uparrow$} \\

\midrule
\multicolumn{12}{c}{\cellcolor{gray!10}\textbf{In-Domain (ID) Evaluation}} \\
\midrule
CC-Edit
& \best{+18.16}
& -78.85
& -6.97
& -4.46
& -20.82
& \best{+5.09}
& \best{+2.94}
& -65.95
& -34.39
& -2.01
& -8.08 \\
UniVideo
& -54.30
& -22.84
& -13.02
& -5.51
& -21.39
& \second{+2.82}
& +0.92
& -25.40
& \second{+2.73}
& \best{+0.82}
& -12.57 \\
SyntheticPip
& -- & -- & -- & -- & -- & -- & -- & -- & -- & -- & -- \\
\textbf{Ours}
& \second{+14.44}
& \best{+0.58}
& -1.34
& \best{+0.16}
& -0.70
& +1.70
& \second{+0.94}
& \best{+20.15}
& \best{+7.46}
& -1.49
& -4.57 \\

\midrule
\multicolumn{12}{c}{\cellcolor{gray!10}\textbf{Out-of-Distribution (OOD) Evaluation}} \\
\midrule
CC-Edit
& -1.74
& -83.40
& \best{+0.22}
& \best{+2.69}
& -16.58
& \best{+5.46}
& \best{+3.21}
& -50.79
& -32.00
& -2.03
& -5.99 \\
UniVideo
& -92.47
& -22.69
& -11.15
& -1.39
& -17.88
& \second{+1.65}
& +0.30
& -29.19
& \second{+3.13}
& \best{+0.36}
& \best{+4.33} \\
SyntheticPip
& -- & -- & -- & -- & -- & -- & -- & -- & -- & -- & -- \\
\textbf{Ours}
& \best{+9.20}
& \best{+1.47}
& -4.52
& -0.22
& -1.25
& +1.19
& \second{+0.77}
& \best{+7.43}
& \best{+8.28}
& \second{+0.13}
& -3.10 \\

\bottomrule
\end{tabular}
\label{tab:main-results-relative-improvement}
\end{table*}

We build upon the video world model Cosmos-Transfer2.5~\citep{nvidia2026worldsimulationvideofoundation}. 
Reference image frames are generated using Nano Banana~\citep{google_nanobanana}. 
Additional details are in App.~\ref{app:metrics}.

In Sec~\ref{sec:exp-main-results} we show that
our model edits driving videos consistently, preserving scene semantics per instruction.
We evaluate on both withheld videos from our in-distribution datasets (Boreas~\citep{burnett_ijrr23}, nuScenes~\citep{caesar2020nuscenes} and Waymo~\citep{sun2020scalability})
and the out-of-distribution nuPlan~\citep{karnchanachari2024towards} dataset.
We explicitly select nuPlan test images that are non-overlapping with any nuScenes training geographic locations. 
Sec.~\ref{sec:downstream} shows the importance of preserving traffic semantics for the downstream lane detection.

\subsection{Main Results on Driving-Scene Video Editing}\label{sec:exp-main-results}
The careful design of our method across data (Sec.~\ref{sec:data}), training (Sec.~\ref{sec:rewards}) and model (Sec.~\ref{sec:image-robustness}) jointly leads to state-of-the-art
traffic-relevant content preservation, particularly lane markings and vehicle appearance, and temporal consistency, while maintaining competitive instruction alignment and photorealism. 

In Tab.~\ref{tab:main-results-relative-improvement}, we compare \ourmethod against the state-of-the-art editors CCEdit~\cite{feng2024ccedit} and UniVideo~\cite{wei2025univideo}, as well as our synthetic generation pipeline (SyntheticPip; Sec.~\ref{sec:data-pseudo-gt}). To facilitate comparison across metrics with heterogeneous scales, we report each method's relative change with respect to SyntheticPip, the most competitive baseline. Raw scores with standard deviations are provided in Tab.~\ref{tab:main-results-raw} in the appendix.

Our framework emphasizes safety-critical structural fidelity. On in-domain data, our model improves temporal consistency given by FF-$\alpha$, and improves content preservation with the best scores for safety-critical road markings and car appearance preservation.
``Ours'' is the only model able to maintain this improvement on out-of-distribution data, highlighting the generalizability of our methods.
Alongside these structural gains, our method maintains competitive instruction alignment and image quality although SyntheticPip scores higher on some metrics in these dimensions.
These quantitative gains are further strongly supported by qualitative analysis. As shown in Fig.~\ref{fig:exp-main-qualitative-results}, our approach successfully preserves safety-critical semantics such as specific vehicle appearances, road lines, crosswalks, and text markings. Baselines like SyntheticPip struggle with hallucinated traffic lights and disappearing vehicles.
We show additional results in Section~\ref{app:additional-results}.

Furthermore, our user study (Tab.~\ref{tab:user_study_pref}) corroborates these findings.
We polled 44 participants, each presented with 10 questions showing video outputs from each method on test examples randomly chosen from our combined In-Domain and Out-of-Distribution evaluation datasets.
For each question, participants evaluated the outputs according to one randomly selected criterion: content preservation, instruction following, photorealism, or temporal consistency (see App.~\ref{app:user-study-details} for details).
The study yielded 429 criterion-level responses, with overall preference computed by pooling responses across criteria.
Users preferred our model across all evaluated criteria, with relative increases in preference of approximately 39\%  in content preservation, 9\% in instruction following, 30\% in photorealism, and 12\% in temporal consistency relative to SyntheticPip.
These results suggest that human viewers perceive our edits as faithful to the prompt while better preserving the underlying scene content.

\begin{table}[t]
\caption{\textbf{User Study.} Preference rates across four evaluation criteria and overall preference pooled across criteria.}
\label{tab:user_study_pref}
\centering
\setlength{\tabcolsep}{3pt}
\renewcommand{\arraystretch}{1.1}
\begin{tabular*}{0.85\linewidth}{@{\extracolsep{\fill}}lccccc@{}}
\toprule
\textbf{Model}
& \shortstack{\textbf{Content}\\\textbf{Preservation}}
& \shortstack{\textbf{Instruction}\\\textbf{Following}}
& \textbf{Photorealism}
& \shortstack{\textbf{Temporal}\\\textbf{Consistency}}
& \textbf{Overall} \\
\midrule
Ours         & \textbf{46\%} & \textbf{36\%} & \textbf{43\%} & \textbf{38\%} & \textbf{40\%} \\
SyntheticPip & 33\% & 33\% & 33\% & 34\% & 33\% \\
UniVideo     & 7\%  & 7\%  & 7\%  & 7\%  & 7\%  \\
CCEdit       & 5\%  & 5\%  & 6\%  & 3\%  & 5\%  \\
Tie          & 8\%  & 20\% & 11\% & 19\% & 15\% \\
\bottomrule
\end{tabular*}
\end{table}

\begin{figure}[ht]
    \centering
    \includegraphics[width=0.9\textwidth]{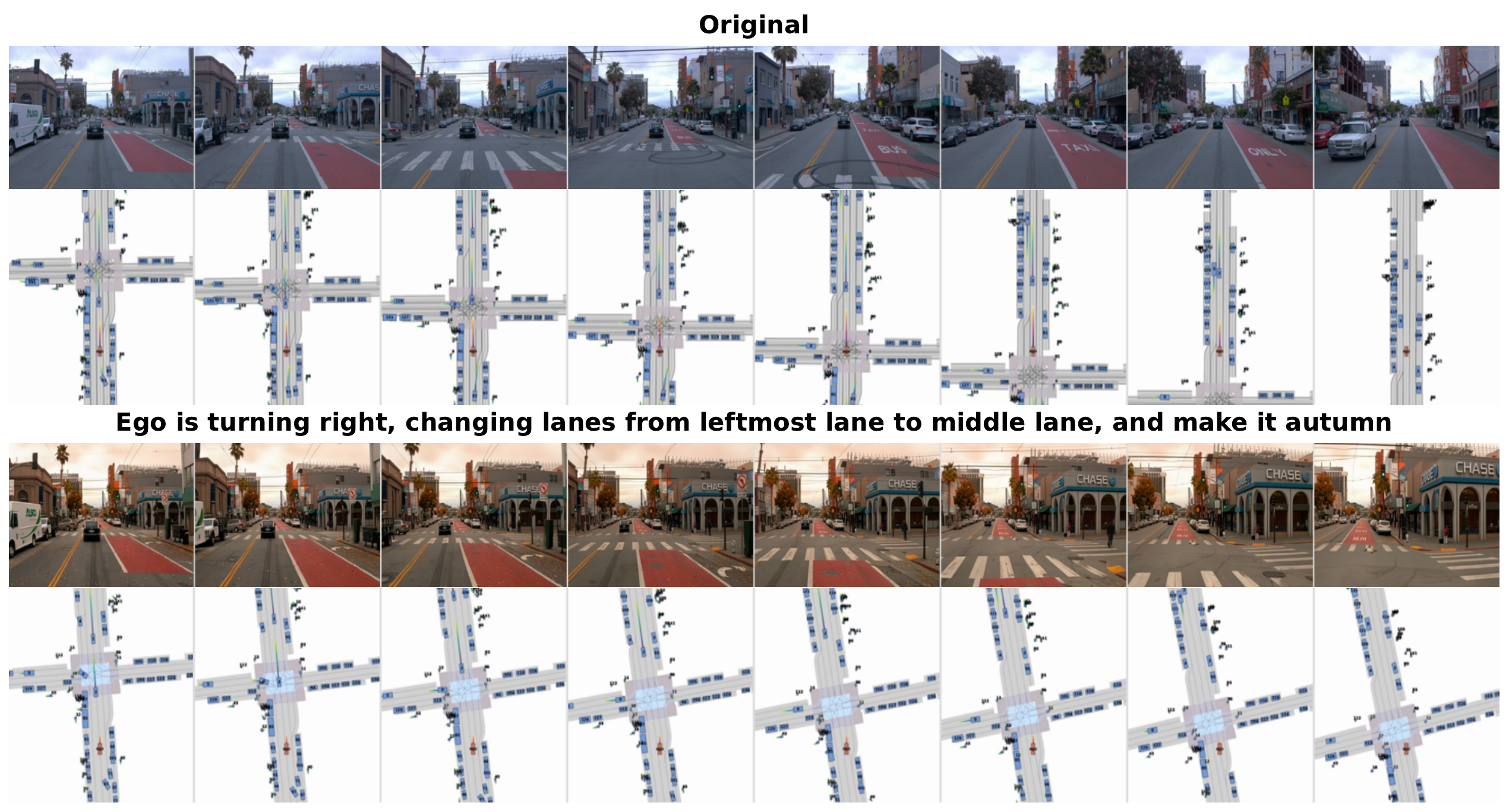}
    \caption{\textbf{Combining Local and Global Appearance Edits.} We apply HorizonComposer on top of HorizonForge~\citep{wang2026horizonforge} to combine an ego-motion edit with a global appearance edit. 
    Given an initial scene, HorizonForge modifies the trajectory to synthesize a new vehicle maneuver (e.g., a right-turn lane change) and generates the corresponding base scene. 
    HorizonComposer then applies global appearance edits--shifting the season to autumn while preserving the newly edited BEV trajectory and local semantics.}
    
    \label{fig:localGlobal}
\end{figure}

\subsection{Extensions to Local Editing}\label{sec:local-editing}

While HorizonComposer is primarily designed for global driving-condition editing, its ability to preserve traffic semantics also makes it readily composable with localized editing approaches. To explore this extension, we build on HorizonForge~\citep{wang2026horizonforge,he2025langdrivectrl,chang2025langtraj}, which offers local trajectory-level controls for modifying driving scenes. Given a source video together with a local editing instruction, for instance, altering the ego-vehicle motion or executing a lane change, HorizonForge first generates a locally edited video that reflects the intended structural modification. HorizonComposer is then applied to this intermediate result to impose the target global appearance condition while retaining the newly synthesized traffic layout. As illustrated in Figure~\ref{fig:localGlobal}, this pipeline achieves high-fidelity global-local editing that jointly captures trajectory adjustments and seasonal transformations. Additional video examples are provided in the supplementary material.

\subsection{Downstream Task: Lane Marking Segmentation}
\label{sec:downstream}

\begin{wraptable}{r}{0.4\linewidth}
\centering
\caption{Lane marking segmentation performance on SeRM.}
\label{tab:serm_seg}
\resizebox{\linewidth}{!}{
\begin{tabular}{lccc}
\toprule
Method & mIoU & Pixel Acc & mAcc \\
\midrule
Original & 37.48 & 98.76 & 50.04 \\
+ SyntheticPip & 38.45 & 98.81 & 53.05 \\
+ Ours & \textbf{40.27} & \textbf{98.88} & \textbf{55.62} \\
\bottomrule
\end{tabular}
}
\end{wraptable}
We evaluate whether our traffic-semantic-preserving video edits benefit downstream perception tasks. Specifically, we conduct lane marking segmentation on the SeRM dataset~\citep{jang2021JAS, jang2018iv}, which provides pixel-level annotations for 16 categories of lane markings, covering diverse road marking types critical for autonomous driving.

We reconstruct full-resolution image–annotation pairs from SeRM and generate additional training data using both the proposed \ourmethod and our best baseline SyntheticPip.
Implementation details are provided in App.~\ref{app:downstream-details}.
We fine-tune a PIDNet~\citep{xu2023pidnet} on three training sets: (1) original data only, (2) original + SyntheticPip-generated data, and (3) original + our generated data. 

We report mIoU, pixel accuracy, and mean accuracy (mAcc) following PIDNet.
As shown in Tab.~\ref{tab:serm_seg}, synthetic data improves segmentation performance, and our method consistently outperforms SyntheticPip across all metrics. In particular, we improve mIoU from 37.48\% to 40.27\% and mAcc from 50.04\% to 55.62\%. The larger gain in mAcc indicates that our method better preserves fine-grained traffic semantics, leading to more effective supervision for safety-critical structures such as road markings.

\subsection{Downstream Task: Multi-Object Detection}
\label{sec:downstream_tracking}

\begin{wraptable}{r}{0.48\linewidth}
\centering
\caption{Downstream multi-object detection performance on BDD100K.}
\label{tab:bdd_detection}
\resizebox{\linewidth}{!}{
\begin{tabular}{lccc}
\toprule
Training Data & mAP $\uparrow$ & mAP@50 $\uparrow$ & mAP@75 $\uparrow$ \\
\midrule
Real-200 & 0.167 & 0.316 & 0.154 \\
Real-200 + Generated-200 & \textbf{0.175} & \textbf{0.324} & \textbf{0.165} \\
\midrule
Relative improvement & +4.79\% & +2.53\% & +7.14\% \\
\bottomrule
\end{tabular}
}
\end{wraptable}

We evaluate whether videos generated by \ourmethod benefit downstream object detection on BDD100K. We compare two fine-tuning settings initialized from the same pretrained checkpoint: (1) fine-tuning on 200 real BDD100K videos, and (2) fine-tuning on the same 200 real videos augmented with 200 additional videos generated by \ourmethod.

We report mean average precision (mAP), together with mAP at IoU thresholds of 0.5 (mAP@50) and 0.75 (mAP@75).
As shown in Tab.~\ref{tab:bdd_detection}, adding our generated videos consistently improves performance across all three metrics. Specifically, mAP increases from 0.167 to 0.175, mAP@50 from 0.316 to 0.324, and mAP@75 from 0.154 to 0.165, corresponding to relative improvements of 4.79\%, 2.53\%, and 7.14\%, respectively. These results suggest that videos generated by \ourmethod provide effective data augmentation for downstream perception tasks.
\section{Conclusion}

We present HorizonComposer, a driving-scene video editing framework for instruction-guided appearance editing. Our method distills and preserves traffic semantics through hybrid real/synthetic supervision, dense reward-conditioned offline reinforcement learning, and temporally robust thinking frames.  HorizonComposer achieves strong spatiotemporal consistency while maintaining faithful instruction alignment. Experiments demonstrate improved preservation of safety-critical structures such as lane markings. The edited videos also improve downstream lane marking segmentation performance, highlighting the potential of traffic-semantic-preserving video editing for scalable autonomous driving simulation and data augmentation.

{
    \small
    \bibliographystyle{plainnat}
    % \nocite{*}
    \bibliography{main}
}

\newpage
\appendix

\section{Limitations}
The efficacy of our dense reward guidance approach is inherently bounded by the accuracy of the underlying reward models. If the dense reward evaluator misclassifies structurally unreliable pixels such as for small, distant agents or under heavy occlusion, or due to neural evaluator hallucinations, the diffusion policy receives noisy improvement signals. Future work could explore joint or iterative training of the reward and diffusion models to mitigate this bottleneck.

HorizonComposer is primarily designed for global appearance edits.
Its standalone capability for localized edits is therefore limited.
Nevertheless, as demonstrated in Section~\ref{sec:local-editing}, our framework can be composed with existing editing pipelines to support localized changes.
Extending the framework to directly support localized editing is a direction for future work.

Our results demonstrate that strong ODD preservation and temporal consistency can be achieved while maintaining competitive instruction following and image quality. Our specialized reward signals primarily target key ODD elements, such as lane markings and vehicles. Extending these signals to broader scene classes is a natural direction for future generalization.

\section{Broader Impacts}
This paper presents work whose goal is to advance the field of machine learning. There are many potential societal consequences of our work, none of which we feel must be specifically highlighted here. Positively, our method enables the generation of high-fidelity, spatiotemporally consistent driving edge-cases. This can be used to enrich simulation environments and train safer, more robust autonomous vehicle systems without the physical risks of real-world testing.

\section{Additional Videos}\label{app:additional-videos}

Additional video results are provided in the supplementary materials. The demo website can be accessed by opening index.html in the unzipped supplementary archive.

\section{Additional Results}\label{app:additional-results}

\begin{figure*}[!htbp]
    \centering
    \hfill
    \begin{subfigure}[b]{0.49\textwidth}
        \centering
        \includegraphics[clip, trim=0cm 0cm 0cm 0cm, width=\textwidth]{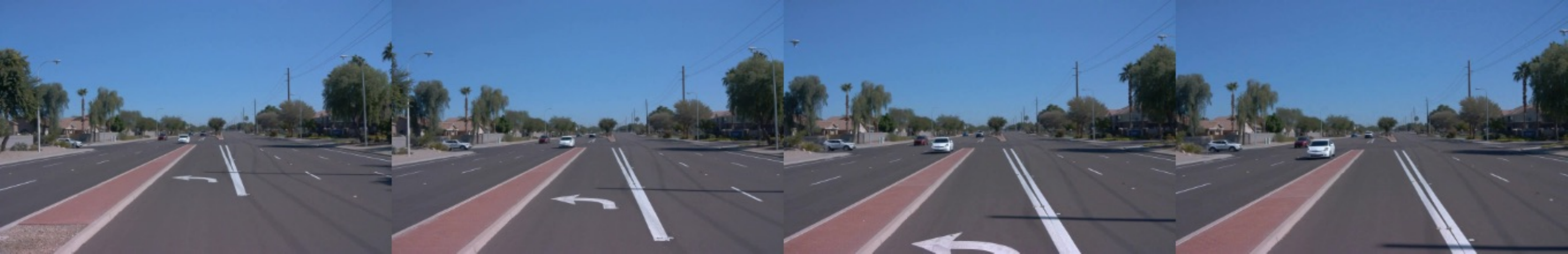}\par
        \includegraphics[clip, trim=0cm 0cm 0cm 0cm, width=\textwidth]{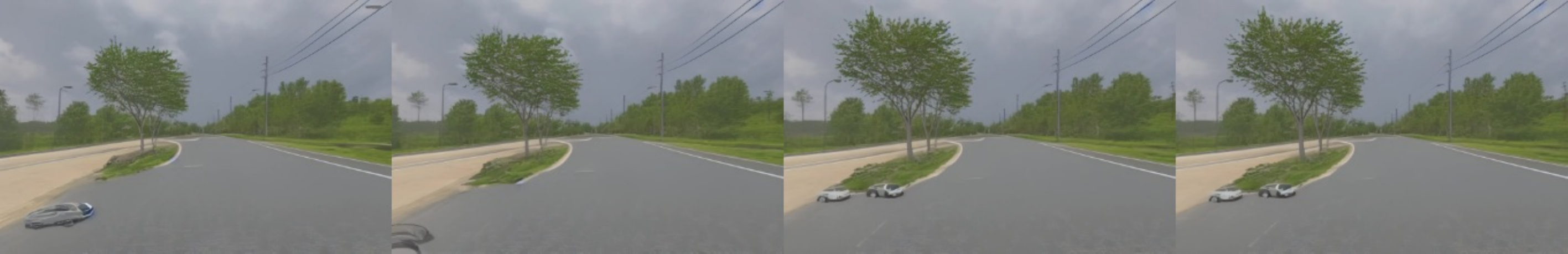}\par
        \includegraphics[clip, trim=0cm 0cm 0cm 0cm, width=\textwidth]{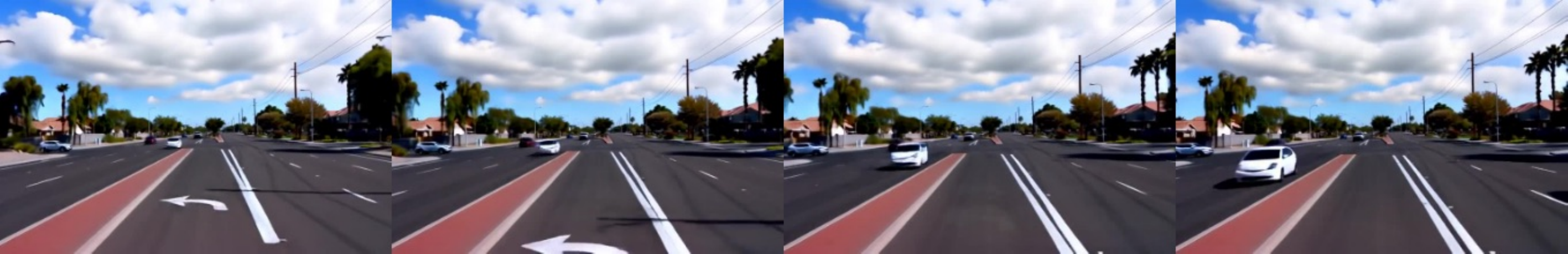}\par
        \includegraphics[clip, trim=0cm 0cm 0cm 0cm, width=\textwidth]{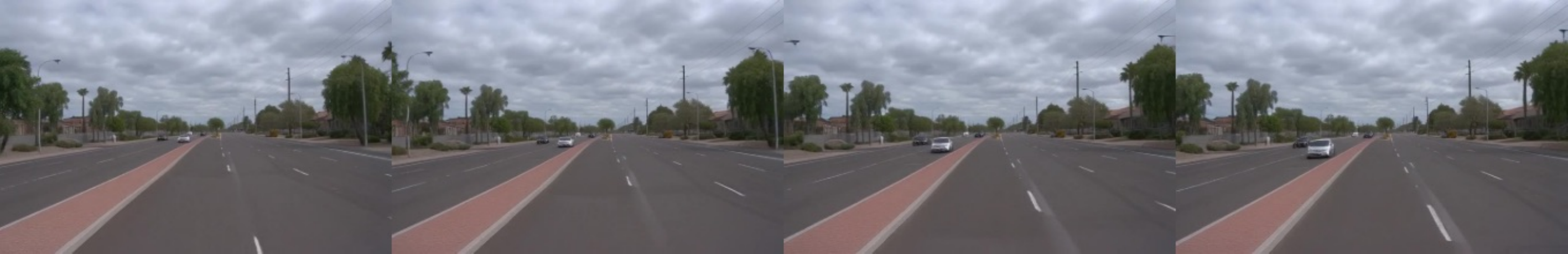}\par
        \includegraphics[clip, trim=0cm 0cm 0cm 0cm, width=\textwidth]{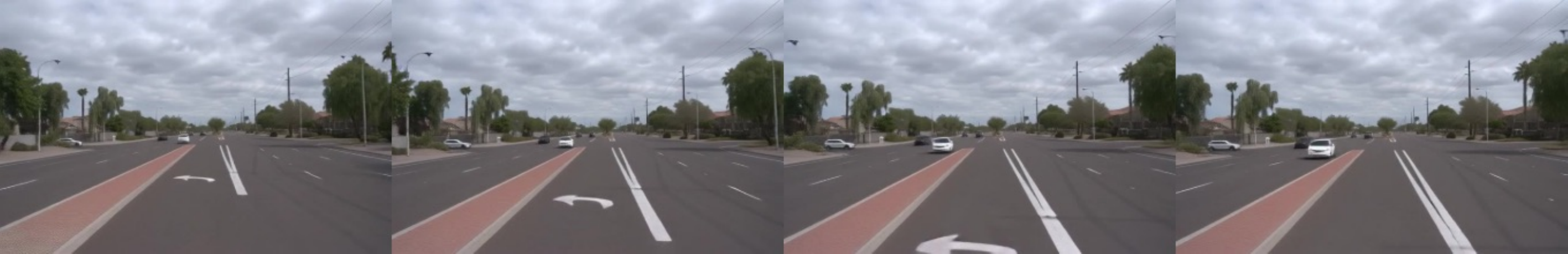}
        \caption{Cloudy conditions during spring.}
    \end{subfigure}
    \hfill
    \begin{subfigure}[b]{0.49\textwidth}
        \centering
        \includegraphics[clip, trim=0cm 0cm 0cm 0cm, width=\textwidth]{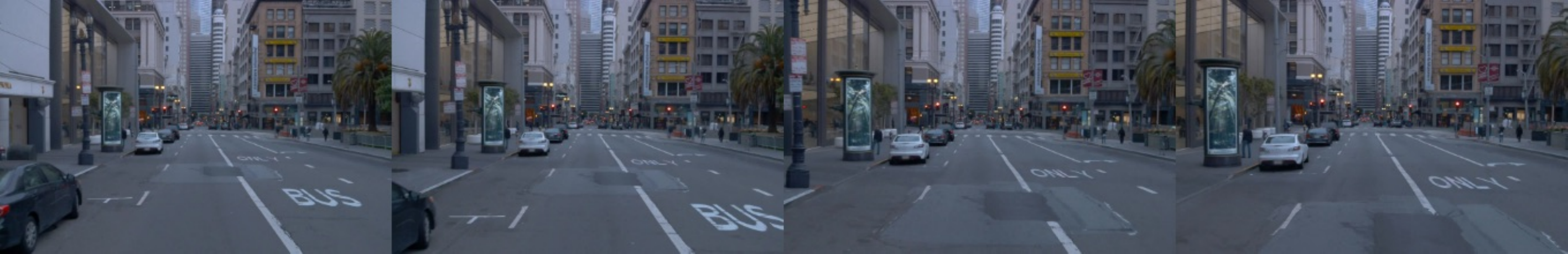}\par
        \includegraphics[clip, trim=0cm 0cm 0cm 0cm, width=\textwidth]{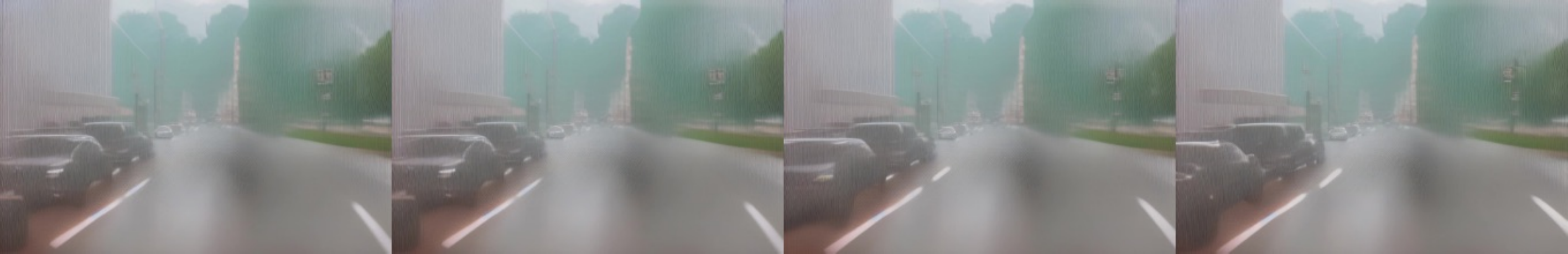}\par
        \includegraphics[clip, trim=0cm 0cm 0cm 0cm, width=\textwidth]{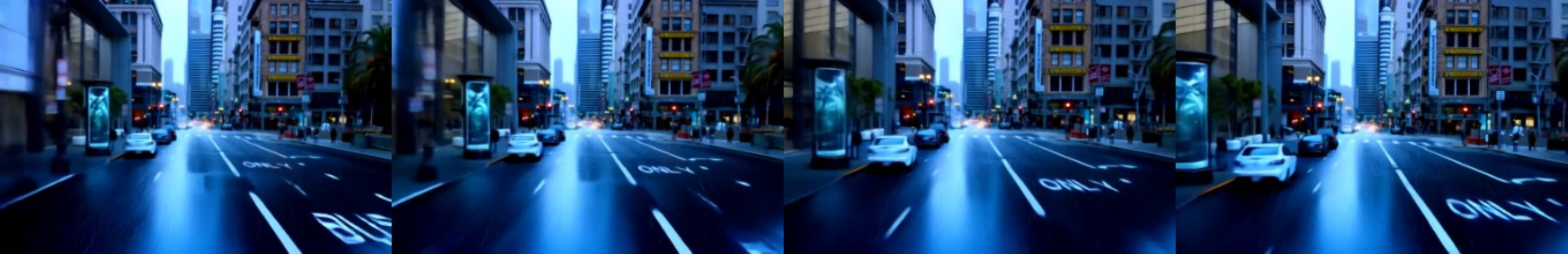}\par
        \includegraphics[clip, trim=0cm 0cm 0cm 0cm, width=\textwidth]{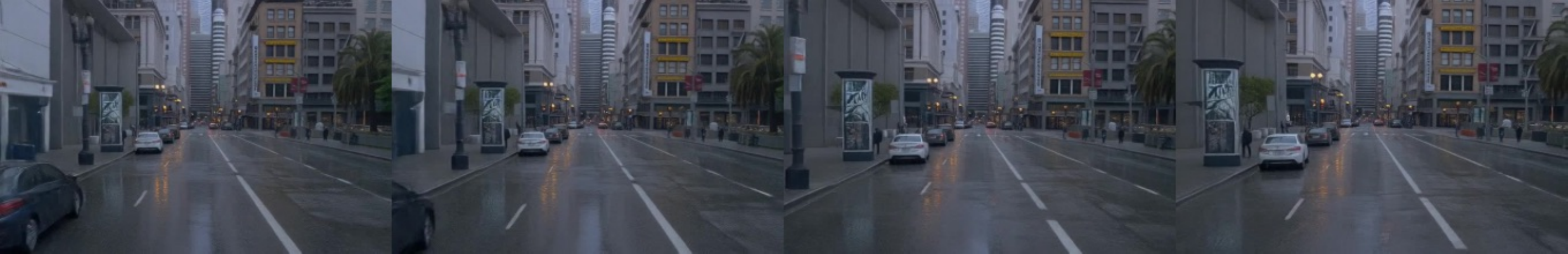}\par
        \includegraphics[clip, trim=0cm 0cm 0cm 0cm, width=\textwidth]{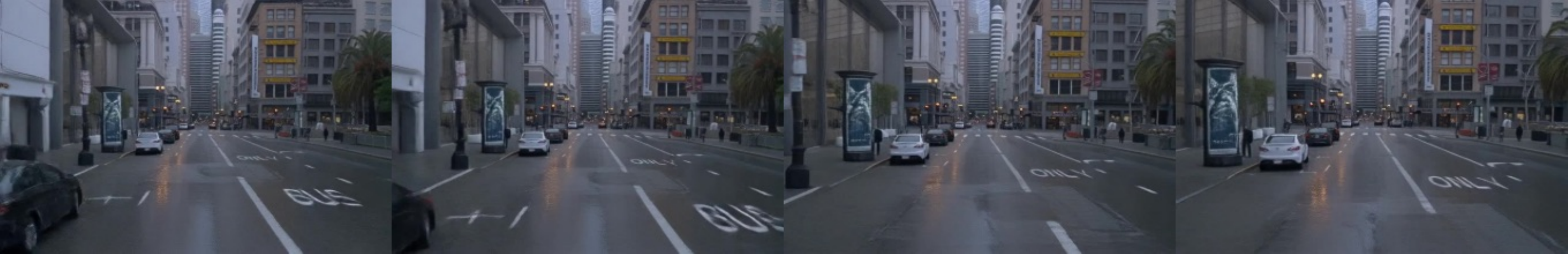}
        \caption{Cloudy conditions during an autumn dusk.}
    \end{subfigure}
    \hfill
    
    \begin{subfigure}[b]{0.49\textwidth}
        \centering
        \includegraphics[clip, trim=0cm 0cm 0cm 0cm, width=\textwidth]{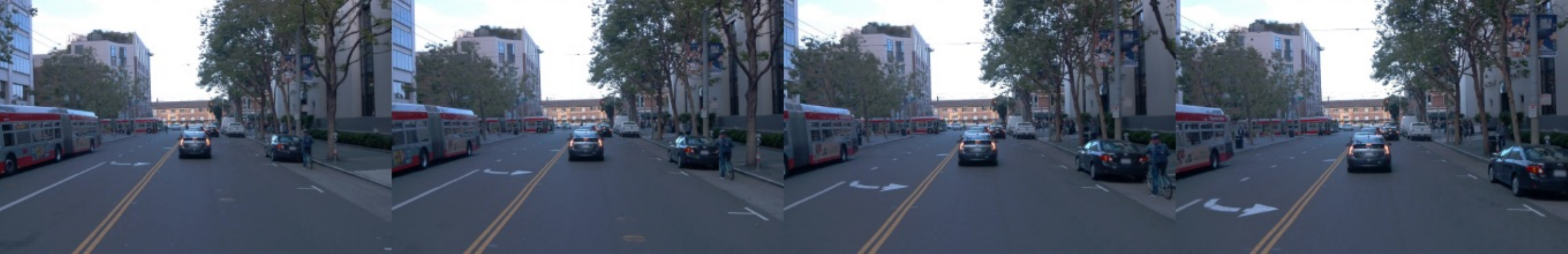}\par
        \includegraphics[clip, trim=0cm 0cm 0cm 0cm, width=\textwidth]{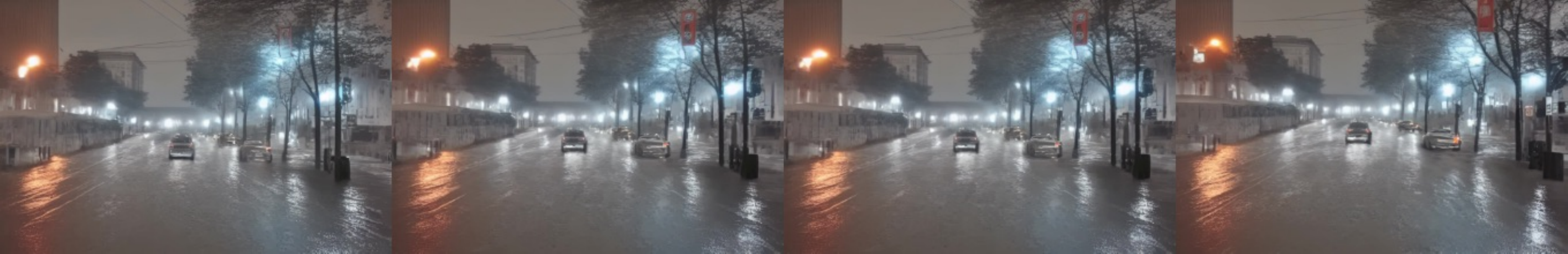}\par
        \includegraphics[clip, trim=0cm 0cm 0cm 0cm, width=\textwidth]{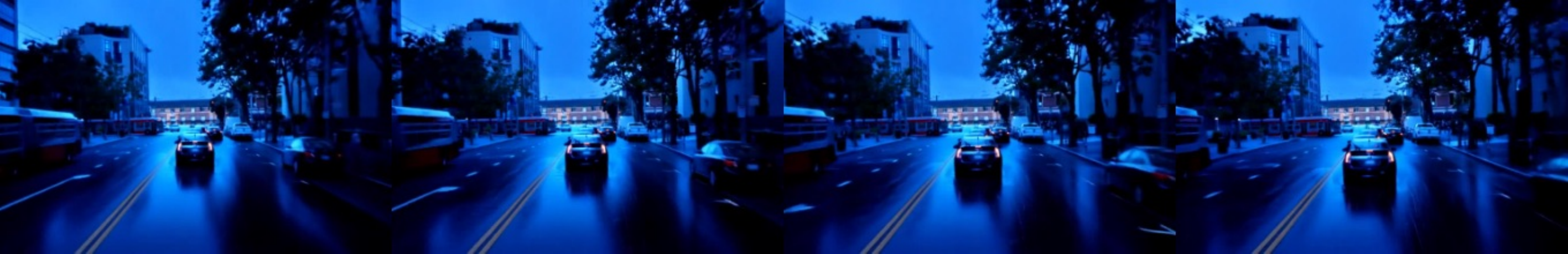}\par
        \includegraphics[clip, trim=0cm 0cm 0cm 0cm, width=\textwidth]{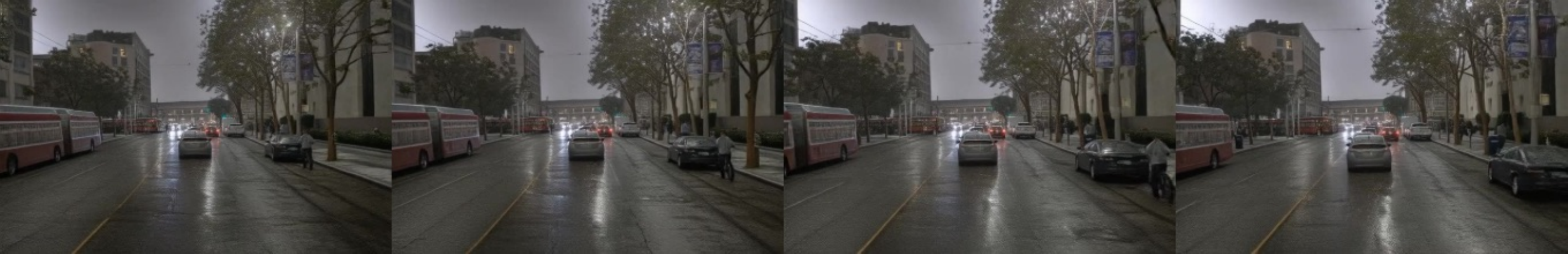}\par
        \includegraphics[clip, trim=0cm 0cm 0cm 0cm, width=\textwidth]{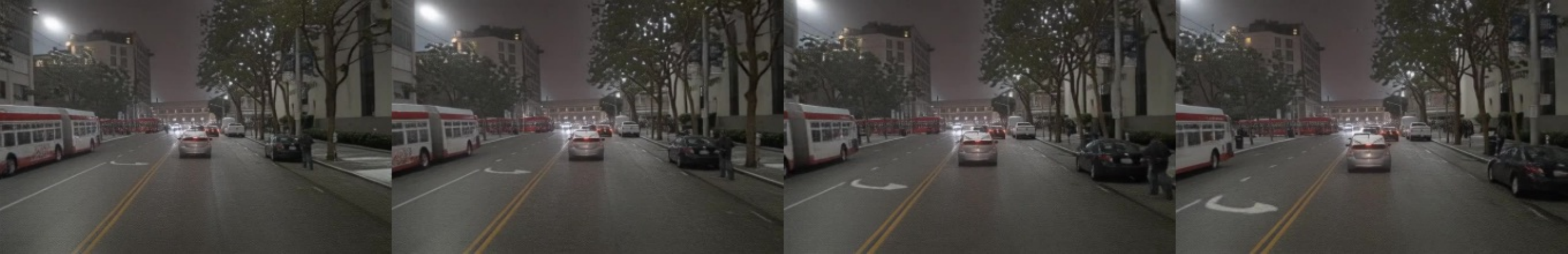}
        \caption{Rainy conditions during a winter night.}
    \end{subfigure}
    \hfill
    \begin{subfigure}[b]{0.49\textwidth}
        \centering
        \includegraphics[clip, trim=0cm 0cm 0cm 0cm, width=\textwidth]{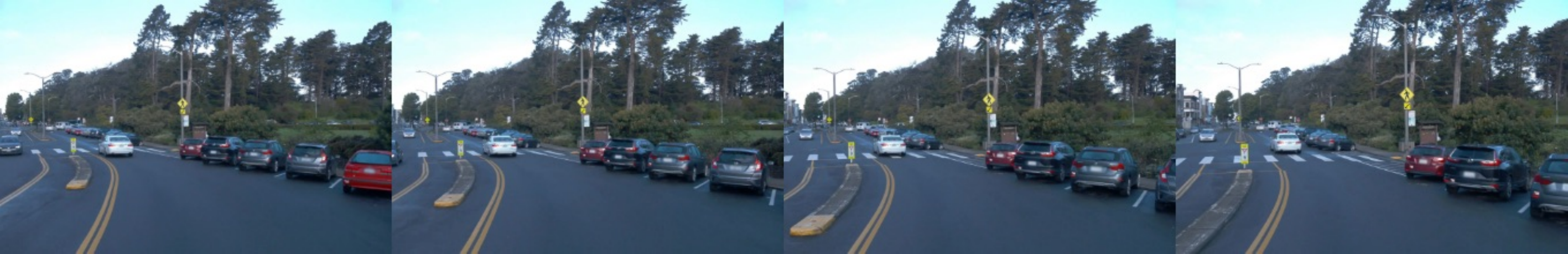}\par
        \includegraphics[clip, trim=0cm 0cm 0cm 0cm, width=\textwidth]{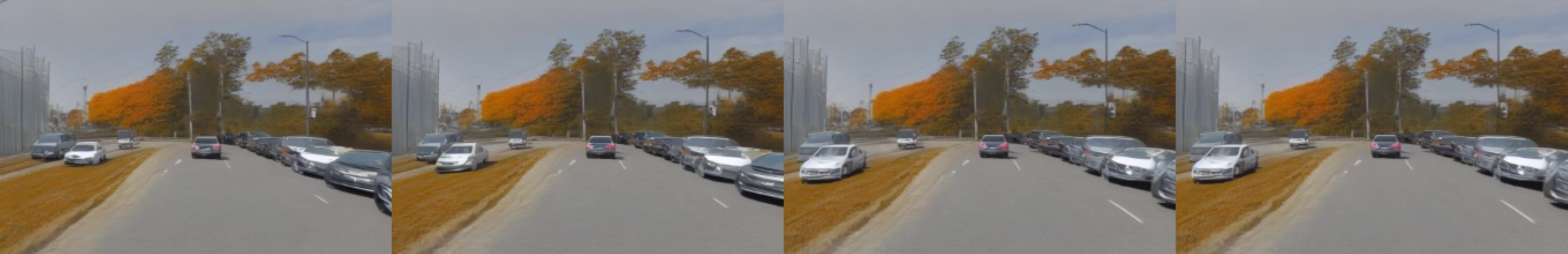}\par
        \includegraphics[clip, trim=0cm 0cm 0cm 0cm, width=\textwidth]{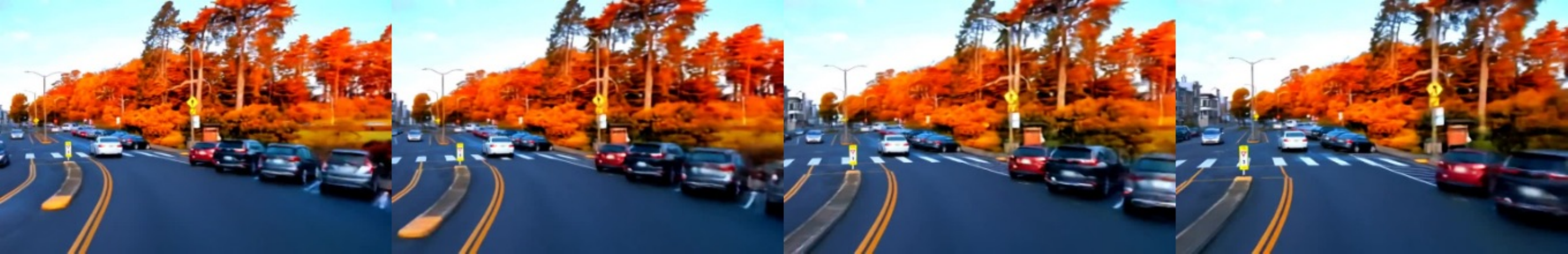}\par
        \includegraphics[clip, trim=0cm 0cm 0cm 0cm, width=\textwidth]{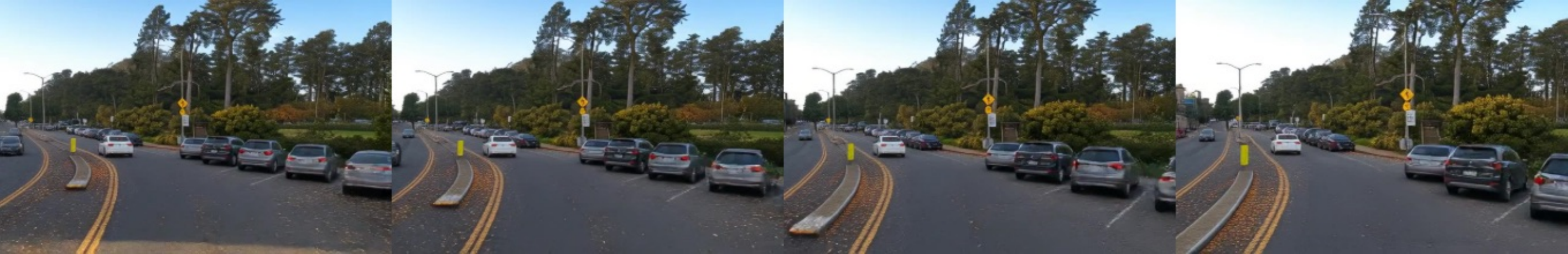}\par
        \includegraphics[clip, trim=0cm 0cm 0cm 0cm, width=\textwidth]{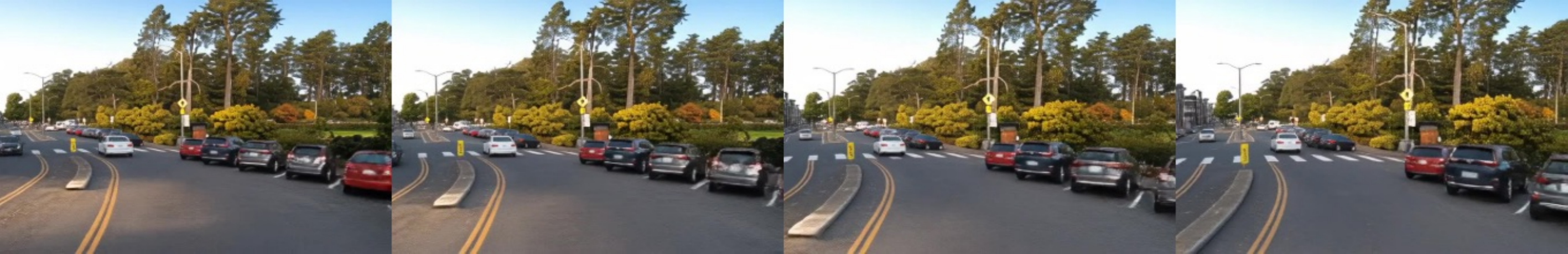}
        \caption{Sunny conditions during a summer dusk.}
    \end{subfigure}
    \hfill

    \begin{subfigure}[b]{0.49\textwidth}
        \centering
        \includegraphics[clip, trim=0cm 0cm 0cm 0cm, width=\textwidth]{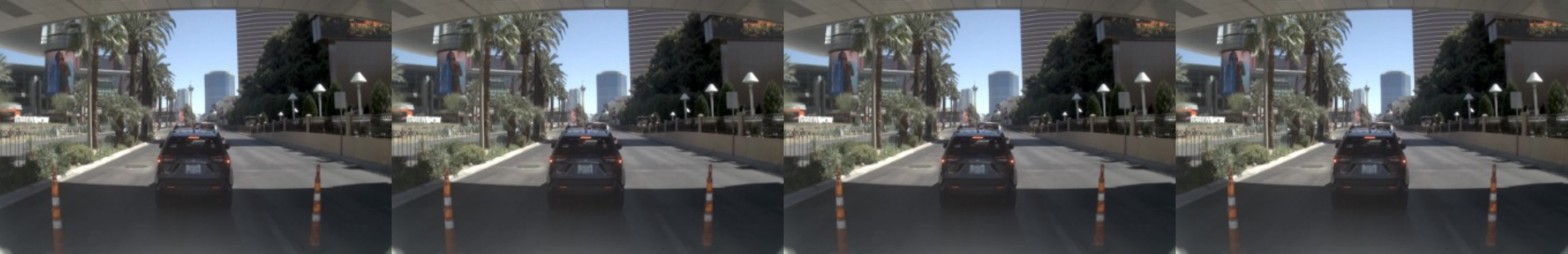}\par
        \includegraphics[clip, trim=0cm 0cm 0cm 0cm, width=\textwidth]{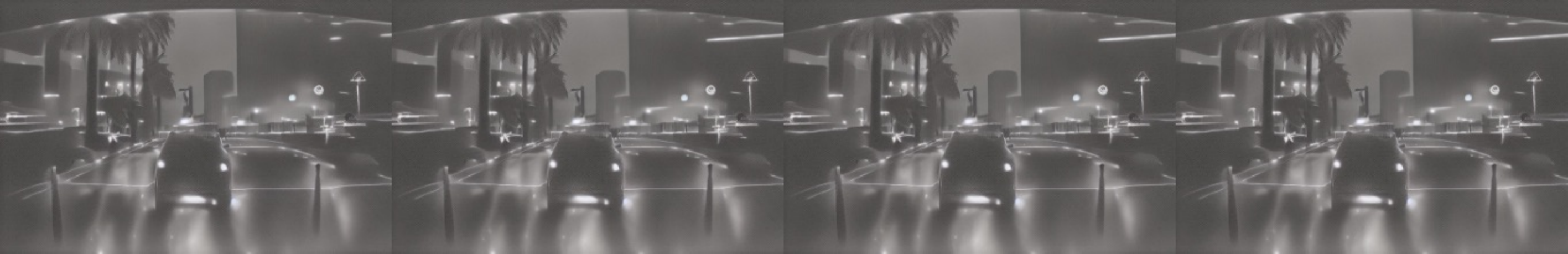}\par
        \includegraphics[clip, trim=0cm 0cm 0cm 0cm, width=\textwidth]{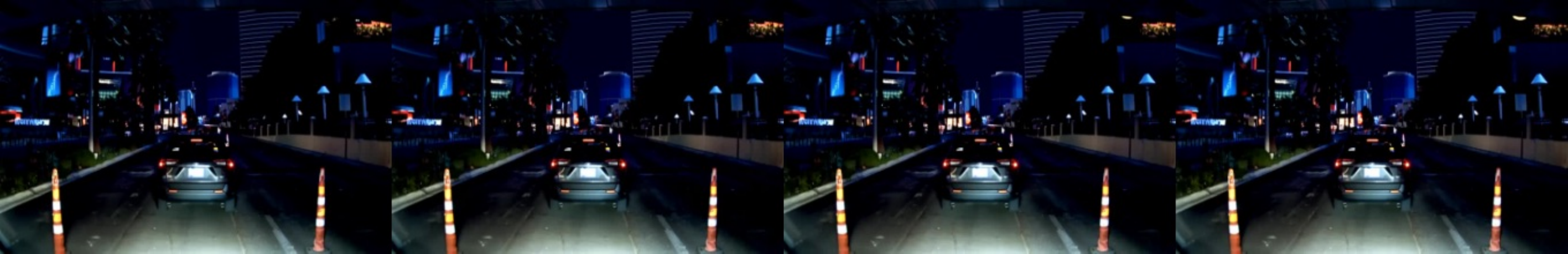}\par
        \includegraphics[clip, trim=0cm 0cm 0cm 0cm, width=\textwidth]{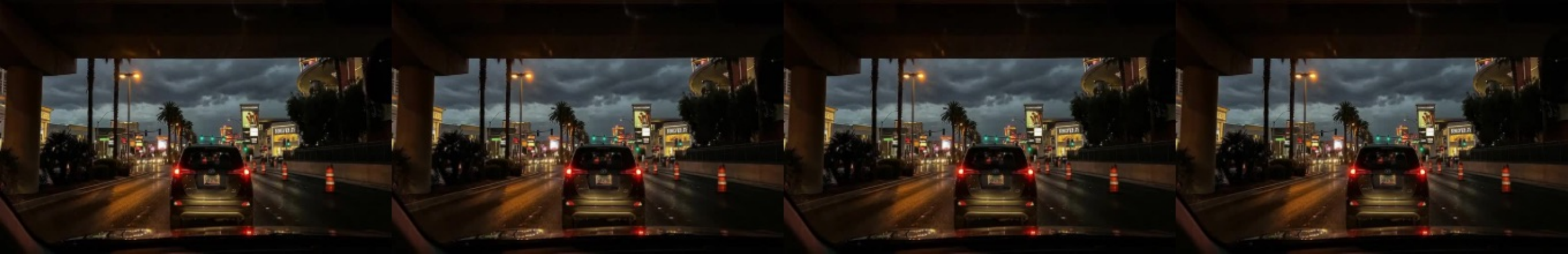}\par
        \includegraphics[clip, trim=0cm 0cm 0cm 0cm, width=\textwidth]{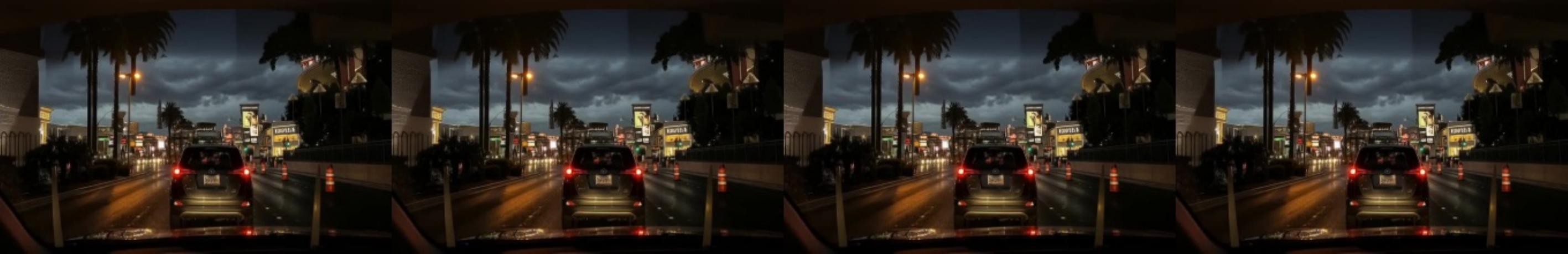}
        \caption{Cloudy conditions at night.}
    \end{subfigure}
    \hfill
    \begin{subfigure}[b]{0.49\textwidth}
        \centering
        \includegraphics[clip, trim=0cm 0cm 0cm 0cm, width=\textwidth]{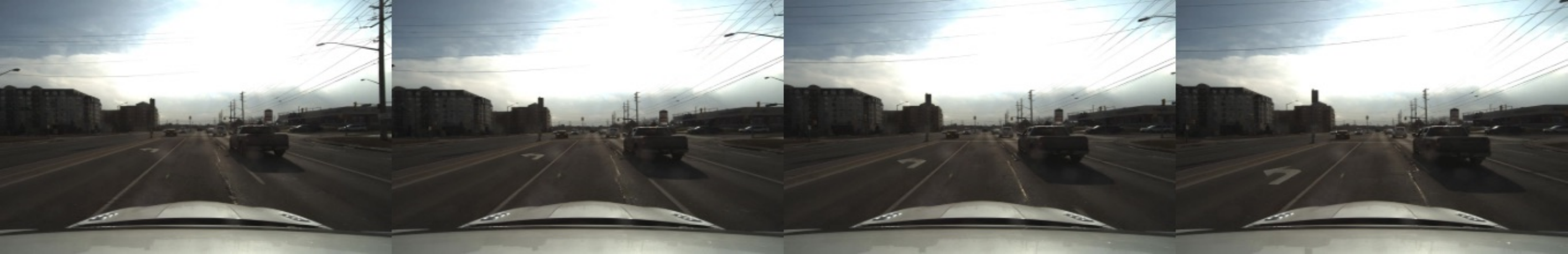}\par
        \includegraphics[clip, trim=0cm 0cm 0cm 0cm, width=\textwidth]{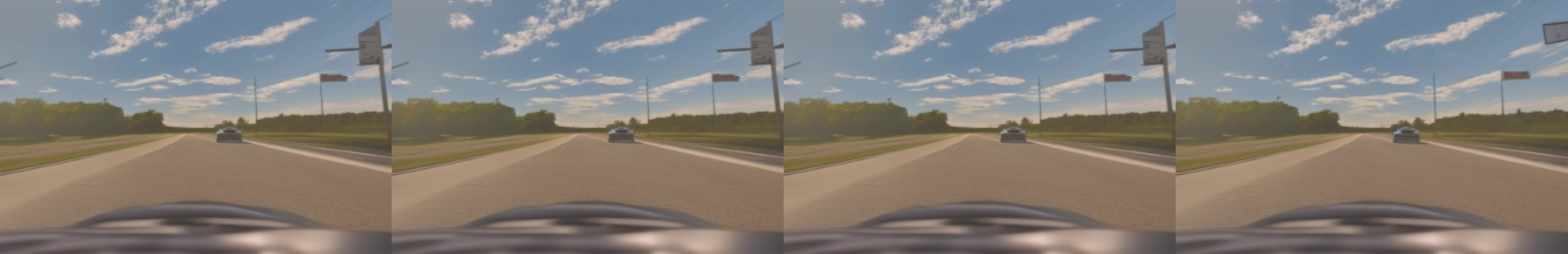}\par
        \includegraphics[clip, trim=0cm 0cm 0cm 0cm, width=\textwidth]{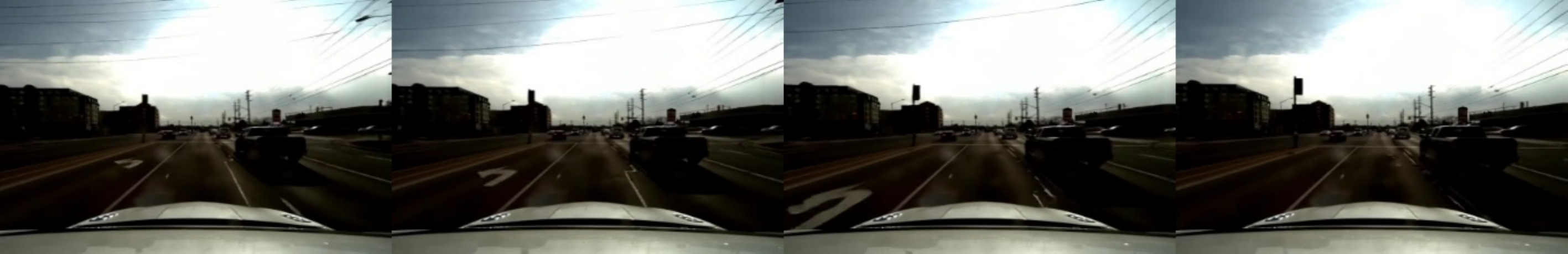}\par
        \includegraphics[clip, trim=0cm 0cm 0cm 0cm, width=\textwidth]{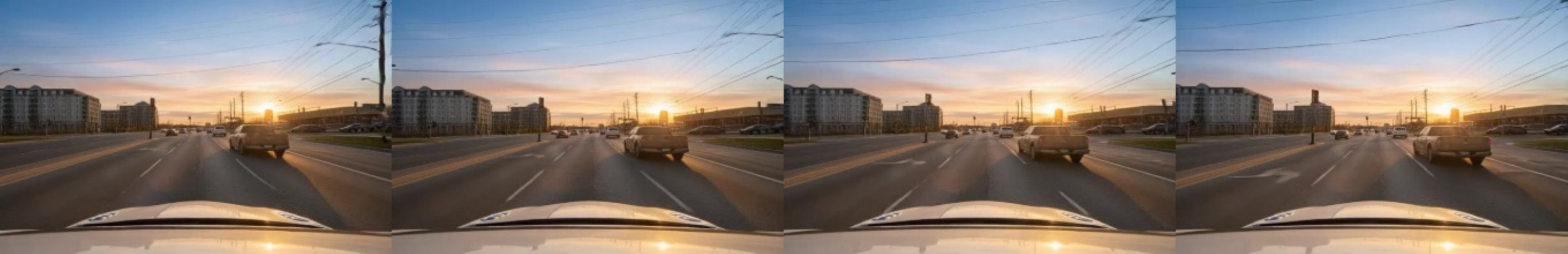}\par
        \includegraphics[clip, trim=0cm 0cm 0cm 0cm, width=\textwidth]{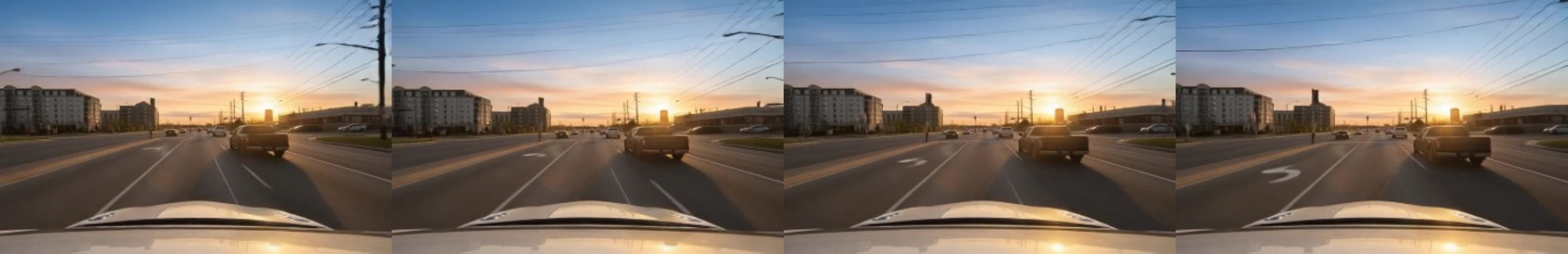}
        \caption{Cloudy conditions during a winter daytime.}
    \end{subfigure}
    \hfill
    
    \caption{\textbf{Qualitative Comparison of Driving Scene Editing.} We highlight key differences in semantic preservation and visual quality across methods. Target conditions are shown in text. Row order from top to bottom: input video, CCEdit, UniVideo, SyntheticPip, Ours. \textbf{(a)} SyntheticPip cannot preserve left turn lane markings. \textbf{(b)} Ours preserves the ``BUS ONLY'' road marking that SyntheticPip removes. \textbf{(c)} Our method successfully preserves the road turn signal that SyntheticPip erroneously removes. \textbf{(d)} Our method successfully preserves the crosswalk and the red car, which SyntheticPip erroneously removes. \textbf{(e)} Our method preserves the clearance height of the overpass. \textbf{(f)} Only UniVideo and Ours preserve the solid white line and turn signal. Overall, CCEdit regularly suffers from hazy artifacts and UniVideo distorts visual realism, whereas our approach maintains high-fidelity instruction alignment and strict Operational Design Domain (ODD) preservation.}
    \label{fig:app-additional-qualitative-results}
\end{figure*}

In this section we demonstrate additional qualitative results in Fig.~\ref{fig:app-additional-qualitative-results} and the raw quantitative results in Tab.~\ref{tab:main-results-raw}.

\begin{table*}[h]
\caption{\textbf{Quantitative Comparison on Driving Scene Editing.}
We evaluate models across temporal stability, instruction alignment,
ODD preservation, and overall video quality. Values indicate means
with standard deviations below. We report \bestinline{best} and
\secondinline{runner-up} mean performances on In-Domain (ID) and
Out-of-Distribution (OOD) scenarios.}
\centering
\footnotesize
\setlength{\tabcolsep}{2.2pt}
\renewcommand{\arraystretch}{0.7}
\begin{tabular}{l cc ccc cccc cc}
\toprule
& \multicolumn{2}{c}{\textbf{Temporal}}
& \multicolumn{3}{c}{\textbf{Edit Align.}}
& \multicolumn{4}{c}{\textbf{ODD Preservation}}
& \multicolumn{2}{c}{\textbf{Quality}} \\
\cmidrule(lr){2-3}
\cmidrule(lr){4-6}
\cmidrule(lr){7-10}
\cmidrule(lr){11-12}
\textbf{Model}
& \textbf{FF-$\alpha\downarrow$}
& \textbf{FF-$\beta\downarrow$}
& \textbf{Succ.$\uparrow$}
& \textbf{CLIP$\uparrow$}
& \textbf{GPT$\uparrow$}
& \textbf{Subj.$\uparrow$}
& \textbf{Bg.$\uparrow$}
& \textbf{Lane$\uparrow$}
& \textbf{Car$\uparrow$}
& \textbf{Aes.$\uparrow$}
& \textbf{Img.$\uparrow$} \\

\midrule
\multicolumn{12}{c}{\cellcolor{gray!10}\textbf{In-Domain Evaluation (Boreas, nuScenes, Waymo)}} \\
\midrule
CC-Edit
& \best{\meanstd{3.2811}{0.0238}}
& \meanstd{0.9638}{0.0005}
& \meanstd{0.8685}{0.0128}
& \meanstd{0.2376}{0.0005}
& \meanstd{5.9436}{0.1320}
& \best{\meanstd{0.9454}{0.0002}}
& \best{\meanstd{0.9710}{0.0000}}
& \meanstd{0.1374}{0.0129}
& \meanstd{0.5103}{0.0118}
& \meanstd{0.4928}{0.0075}
& \meanstd{0.5667}{0.0030} \\[3pt]
UniVideo
& \meanstd{6.186}{0.140}
& \meanstd{0.662}{0.002}
& \meanstd{0.812}{0.013}
& \meanstd{0.235}{0.001}
& \meanstd{5.901}{0.106}
& \second{\meanstd{0.925}{0.001}}
& \meanstd{0.952}{0.001}
& \meanstd{0.301}{0.003}
& \second{\meanstd{0.799}{0.026}}
& \best{\meanstd{0.507}{0.002}}
& \meanstd{0.539}{0.004} \\[3pt]
SyntheticPip
& \meanstd{4.0092}{0.1992}
& \second{\meanstd{0.5389}{0.0038}}
& \best{\meanstd{0.9336}{0.0071}}
& \second{\meanstd{0.2487}{0.0006}}
& \best{\meanstd{7.5063}{0.0380}}
& \meanstd{0.8996}{0.0022}
& \meanstd{0.9433}{0.0006}
& \second{\meanstd{0.4035}{0.0020}}
& \meanstd{0.7778}{0.0070}
& \second{\meanstd{0.5029}{0.0050}}
& \best{\meanstd{0.6165}{0.0089}} \\[3pt]
\textbf{Ours}
& \second{\meanstd{3.4303}{0.0775}}
& \best{\meanstd{0.5358}{0.0065}}
& \second{\meanstd{0.9211}{0.0052}}
& \best{\meanstd{0.2491}{0.0001}}
& \second{\meanstd{7.4541}{0.0219}}
& \meanstd{0.9149}{0.0008}
& \second{\meanstd{0.9522}{0.0003}}
& \best{\meanstd{0.4848}{0.0030}}
& \best{\meanstd{0.8358}{0.0037}}
& \meanstd{0.4954}{0.0025}
& \second{\meanstd{0.5883}{0.0016}} \\

\midrule
\multicolumn{12}{c}{\cellcolor{gray!10}\textbf{Out-of-Distribution Evaluation (nuPlan)}} \\
\midrule
CC-Edit
& \meanstd{3.7547}{0.0607}
& \meanstd{0.9612}{0.0004}
& \best{\meanstd{0.9294}{0.0114}}
& \best{\meanstd{0.2364}{0.0003}}
& \meanstd{6.2439}{0.1059}
& \best{\meanstd{0.9503}{0.0006}}
& \best{\meanstd{0.9755}{0.0012}}
& \meanstd{0.2849}{0.0122}
& \meanstd{0.5387}{0.0161}
& \meanstd{0.5408}{0.0001}
& \meanstd{0.5884}{0.0039} \\[3pt]
UniVideo
& \meanstd{7.103}{0.135}
& \meanstd{0.643}{0.002}
& \meanstd{0.824}{0.008}
& \meanstd{0.227}{0.002}
& \meanstd{6.146}{0.066}
& \second{\meanstd{0.916}{0.002}}
& \meanstd{0.948}{0.002}
& \meanstd{0.410}{0.005}
& \second{\meanstd{0.817}{0.002}}
& \best{\meanstd{0.554}{0.004}}
& \best{\meanstd{0.653}{0.009}} \\[3pt]
SyntheticPip
& \second{\meanstd{3.6904}{0.3114}}
& \second{\meanstd{0.5241}{0.0062}}
& \second{\meanstd{0.9274}{0.0021}}
& \second{\meanstd{0.2302}{0.0009}}
& \best{\meanstd{7.4845}{0.0013}}
& \meanstd{0.9011}{0.0009}
& \meanstd{0.9452}{0.0012}
& \second{\meanstd{0.5790}{0.0023}}
& \meanstd{0.7922}{0.0060}
& \meanstd{0.5520}{0.0079}
& \second{\meanstd{0.6259}{0.0187}} \\[3pt]
\textbf{Ours}
& \best{\meanstd{3.3507}{0.1156}}
& \best{\meanstd{0.5164}{0.0065}}
& \meanstd{0.8855}{0.0082}
& \meanstd{0.2297}{0.0004}
& \second{\meanstd{7.3907}{0.0189}}
& \meanstd{0.9118}{0.0008}
& \second{\meanstd{0.9525}{0.0004}}
& \best{\meanstd{0.6220}{0.0062}}
& \best{\meanstd{0.8578}{0.0019}}
& \second{\meanstd{0.5527}{0.0050}}
& \meanstd{0.6065}{0.0047} \\

\bottomrule
\end{tabular}
\label{tab:main-results-raw}
\end{table*}

\subsection{Ablation: Robust Supervision \& Conditioning}\label{sec:abl-main-components}

\begin{table}[h]
\caption{\textbf{Ablation Study.} Impact of our methods across data (Sec.~\ref{sec:data}), training (Sec.~\ref{sec:rewards}) and model (Sec.~\ref{sec:image-robustness}) on temporal stability and ODD preservation on out-of-distribution data.}
\centering
\footnotesize
\setlength{\tabcolsep}{3.2pt} 
\renewcommand{\arraystretch}{0.9}
\begin{tabular}{l cc ccc cccc cc}
\toprule
& \multicolumn{2}{c}{\textbf{Temporal}} & \multicolumn{3}{c}{\textbf{Edit Align.}} & \multicolumn{4}{c}{\textbf{ODD Preservation}} & \multicolumn{2}{c}{\textbf{Quality}} \\
\cmidrule(lr){2-3} \cmidrule(lr){4-6} \cmidrule(lr){7-10} \cmidrule(lr){11-12}
\textbf{Model Variant} & \textbf{FF-$\alpha$↓} & \textbf{FF-$\beta$↓} & \textbf{Succ.↑} & \textbf{CLIP↑} & \textbf{GPT↑} & \textbf{Subj.↑} & \textbf{Bg.↑} & \textbf{Lane↑} & \textbf{Car↑} & \textbf{Aes.↑} & \textbf{Img.↑} \\
\midrule
SyntheticPip        & 3.4702 & 0.5285 & \best{0.9289} & 0.2308 & \second{7.4855} & 0.9018 & 0.9443 & 0.5774 & 0.7880 & 0.5465 & \best{0.6126} \\
Finetune RGB only & 3.3009 & 0.5505 & \second{0.9234} & \best{0.2353} & 7.4600 & \second{0.9038} & \second{0.9457} & 0.6006 & 0.8243 & 0.5456 & 0.5986 \\
+ Depth Cond.    & \best{3.1225} & 0.5442 & 0.9227 & \second{0.2334} & \best{7.4970} & 0.9013 & 0.9433 & 0.5984 & 0.8308 & 0.5461 & 0.5985 \\
+ Dense Rewards  & 3.3927 & \best{0.5225} & 0.9048 & 0.2307 & 7.4206 & 0.9011 & 0.9436 & \second{0.6266} & \best{0.8595} & \best{0.5482} & \second{0.6066} \\
\textbf{+ Temp. Aug. (Ours)} & \second{3.2185} & \second{0.5235} & 0.8762 & 0.2299 & 7.3836 & \best{0.9126} & \best{0.9525} & \best{0.6281} & \second{0.8558} & \second{0.5472} & 0.6030 \\
\bottomrule
\end{tabular}
\label{tab:ablation_methods_ood}
\end{table}

\begin{table}[h]
\caption{\textbf{Ablation Study (In-Distribution).} Impact of our methods across data (Sec.~\ref{sec:data}), training (Sec.~\ref{sec:rewards}) and model (Sec.~\ref{sec:image-robustness}) on temporal stability and ODD preservation on in-distribution data.}
\centering
\footnotesize
\setlength{\tabcolsep}{3.2pt} 
\renewcommand{\arraystretch}{0.9}
\begin{tabular}{l cc ccc cccc cc}
\toprule
& \multicolumn{2}{c}{\textbf{Temporal}} & \multicolumn{3}{c}{\textbf{Edit Align.}} & \multicolumn{4}{c}{\textbf{ODD Preservation}} & \multicolumn{2}{c}{\textbf{Quality}} \\
\cmidrule(lr){2-3} \cmidrule(lr){4-6} \cmidrule(lr){7-10} \cmidrule(lr){11-12}
\textbf{Model Variant} & \textbf{FF-$\alpha$↓} & \textbf{FF-$\beta$↓} & \textbf{Succ.↑} & \textbf{CLIP↑} & \textbf{GPT↑} & \textbf{Subj.↑} & \textbf{Bg.↑} & \textbf{Lane↑} & \textbf{Car↑} & \textbf{Aes.↑} & \textbf{Img.↑} \\
\midrule
SyntheticPip        & 3.8055 & \second{0.5427} & 0.9375 & 0.2491 & 7.4934 & 0.9004 & 0.9427 & 0.4029 & 0.7859 & 0.4981 & \best{0.6067} \\
Finetune RGB only & 3.6594 & 0.5716 & \best{0.9569} & \best{0.2545} & \best{7.5885} & \second{0.9073} & \second{0.9485} & 0.4133 & 0.7917 & \best{0.5027} & \second{0.5998} \\
+ Depth Cond.    & \second{3.4901} & 0.5585 & \second{0.9489} & \second{0.2525} & \second{7.5848} & 0.9019 & 0.9450 & 0.4276 & 0.8008 & \second{0.5018} & 0.5932 \\
+ Dense Rewards  & 3.6197 & \best{0.5411} & 0.9333 & 0.2500 & 7.4897 & 0.8987 & 0.9426 & \best{0.4904} & \second{0.8262} & 0.4975 & 0.5979 \\
\textbf{+ Temp. Aug. (Ours)} & \best{3.3539} & 0.5428 & 0.9155 & 0.2492 & 7.4297 & \best{0.9156} & \best{0.9520} & \second{0.4845} & \best{0.8351} & 0.4928 & 0.5873 \\
\bottomrule
\end{tabular}
\label{tab:ablation_methods_id}
\end{table}

To understand how our method protects traffic semantics during editing, we ablate our design in Tab.~\ref{tab:ablation_methods_ood} and Tab.~\ref{tab:ablation_methods_id}. Our ablation reveals:
\textbf{1) Standard finetuning is insufficient.} While the SyntheticPip baseline and our minimal RGB-only finetune achieve high Edit Alignment, they suffer from semantic corruption, exhibiting the poorest Lane and Car preservation scores. 
\textbf{2) Rewards protect localized semantics.} Introducing dense reward conditioning yields a massive improvement in ODD Preservation. Specifically, Lane and Car appearance preservation jump significantly across both ID and OOD splits, proving that our localized rewards successfully penalize structural hallucinations.
\textbf{3) Thinking frames stabilize temporal dynamics.} Adding thinking frames (``Ours'') yields the highest Subject and Background consistency while improving intensity-based temporal fidelity (FF-$\alpha\downarrow$). By buffering against temporal shock, our full model strictly maintains foundational scene structure. Crucially, these structural improvements do not compromise the intended edit. While automated metrics suggest a slight trade-off, qualitative and human evaluations (Sec.~\ref{sec:exp-main-results}) confirm that our full method maintains competitive imaging and aesthetic quality, yielding edits that users perceive as highly instruction-aligned while preserving the underlying physical reality.

\section{Detailed Related Works}~\label{app:detailed-related-works}

\mypara{Instruction-Guided Image/Video Editing}
Recent advancements in video generative models, such as \citep{wan2025wanopenadvancedlargescale, hacohen2024ltxvideorealtimevideolatent, kong2025hunyuanvideosystematicframeworklarge}, have significantly improved general-purpose video synthesis. Yet they can fall short in domains requiring precise, fine-grained control over object dynamics and physical consistency. To improve semantic control, a growing body of work focuses on unifying multimodal understanding and generation.
These frameworks typically achieve this by interleaving text, image, and video tokens into a single unified sequence or by tightly integrating Multimodal Large Language Models (MLLMs) with Diffusion Transformers \citep{wei2025univideo, mou2025instructxunifiedvisualediting, jiang2025vaceallinonevideocreation, ju2026editverseunifyingimagevideo, chen2026vinounifiedvisualgenerator}.
Other approaches seek to explicitly decouple structural and visual guidance or employ tuning-free temporal propagation to edit videos while preserving layout \citep{wu2023tuneavideooneshottuningimage, ku2024anyv2vtuningfreeframeworkvideotovideo, feng2024ccedit}. Conceptually, our framework shares the philosophy of decoupling structure and appearance; however, these general-purpose models cannot natively guarantee the strict spatiotemporal traffic semantics required for driving scenes.
Closest to our temporal approach is ChronoEdit \citep{wu2025chronoedittemporalreasoningimage}, which tackles physical consistency in image editing by hallucinating intermediate video frames as "reasoning tokens" to map out a plausible physical trajectory. Our proposed \emph{thinking frames} shares this core philosophy. However, while ChronoEdit applies this to single-image edits, we utilize thinking frames to structurally anchor continuous multi-agent dynamics and insulate the generative pipeline against conditioning-induced temporal shock.

\mypara{Driving Scene \& Global Appearance Synthesis}
The rising demand for scalable driving-scene data has spurred the development of driving scene simulation methods through generation~\citep{swerdlow2024BEVGen,yang2023bevcontrol,wang2023drivedreamer,gao2023magicdrive,wen2024panacea,gao2024magicdrivedit,lu2024infinicube,chen2025autoscape,wang2026horizonforge, soroco2026horizonweaver,ljungbergh2025r3d2}. Models such as the Cosmos family \citep{alhaija2025cosmos,nvidia2026worldsimulationvideofoundation} scale physical prediction and sim-to-real translation into unified frameworks. Yet, relying purely on massive, implicit data priors often fails to strictly preserve the Operational Design Domain (ODD) during targeted appearance edits.
To enforce spatial consistency, many recent approaches incorporate explicit 3D/4D representations or physics-based priors. These include pipelines that combine 2D background editing with 4D weather particle fields \citep{qian2025weathereditcontrollableweatherediting}, hybrid systems that pair element-wise pixel/object controls with physics simulators \citep{mo2025dreamlandcontrollableworldcreation}, unified occupancy-centric frameworks \citep{li2025unisceneunifiedoccupancycentricdriving}, and representations that decompose dynamic scenes into view-dependent neural graphs \citep{schneider2025neuralatlasgraphsdynamic}.
While these methods offer spatial conditioning to manipulate traffic semantics, they rely on external simulation engines, and nonetheless struggle to preserve original vehicle identities or scene structures without hallucinating new structures.
In contrast, our approach operates directly in the video domain. By distilling traffic semantics through dense, pixel-level rewards and a hybrid data pairing strategy, we strictly enforce ODD preservation and achieve spatiotemporal consistency without the overhead of explicit 3D or 4D reconstruction.

\mypara{Reinforcement learning for diffusion models}
Adapting reinforcement learning to diffusion models typically relies on 
policy-gradient methods \citep{liu2025flowgrpotrainingflowmatching, zhao2026imageeditr1boostingmultiagentimage, black2024trainingdiffusionmodelsreinforcement, li2024instructrl4pixtrainingdiffusionimage} (e.g., PPO), which face prohibitive sampling costs for video, or offline weighted regression methods like AWR~\citep{peng2019advantageweightedregressionsimplescalable}, which discard valuable data and degrade into filtered imitation learning~\citep{intelligence2025pi06vlalearnsexperience}. 
Recent work~\citep{jin2025inferencetimealignmentcontroldiffusion, frans2025diffusion, cheng2026diffusionblendinferencetimemultipreference, intelligence2025pi06vlalearnsexperience} addressed these issues by shifting policy improvement to inference via reward-conditioned guidance. 
Yet many rely on expensive auxiliary models or sample-level scalar indicators which fail to capture how structural reliability varies drastically \textit{within} a video frame. We extend this line of work using dense, pixel-aligned reward indicators to extract localized policy improvements from partially reliable hybrid data.

\section{Real-World Dataset Construction} \label{app:real-world-data}

Expanding on Section~\ref{sec:data-real-world}, we provide additional details on the real-world data construction pipeline by walking through each of its modules.

\mypara{Pose-based Pairing} Our pipeline starts with a pairing process that matches frames between two traversals based on spatial proximity and orientation similarity. Concretely, we designate one traversal as the source video and seek, for each source frame, the most pose-similar frame from another traversal of the same route. Formally, for each source frame $I_{\text{source}}$ in the source sequence, we define its camera pose as $(\vec{x}_s, \phi_s, \theta_s, \psi_s)$, where $\vec{x}_s \in \mathbb{R}^3$ is the camera position and $(\phi_s, \theta_s, \psi_s)$ denote roll, pitch, and yaw. We then define a local temporal neighborhood $\mathcal{I}$ of candidate frames drawn from a second traversal of the same route, and select the paired target frame $I_{\text{target}}$ by minimizing a pose-disparity distance following HorizonWeaver~\citep{soroco2026horizonweaver}:
\begin{equation}\label{eqn:min-camera-pose}
\begin{aligned}
I_{\text{target}} &= \arg\min_{I \in \mathcal{I}} \; \text{dist}(I, I_{\text{source}}), \\
\text{dist}(I_a, I_b) &= 
\bigl\lVert \vec{x}_a - \vec{x}_b \bigr\rVert_2
+ \bigl| \phi_a - \phi_b \bigr|
+ \bigl| \theta_a - \theta_b \bigr| \\
&~\quad+ \bigl| \psi_a - \psi_b \bigr|,
\end{aligned}
\end{equation}
where $(\vec{x}_a, \phi_a, \theta_a, \psi_a)$ and $(\vec{x}_b, \phi_b, \theta_b, \psi_b)$ are the poses of frames $I_a$ and $I_b$, respectively. In practice, we accept matches only when $\text{dist}(I_{\text{target}}, I_{\text{source}})$ falls below a fixed threshold, ensuring that paired frames share nearly identical viewpoints. This procedure is dataset-agnostic and applies to any driving dataset with sufficiently accurate pose estimates, including but not limited to Boreas, yielding paired image sequences that depict the same physical locations under different conditions.

\mypara{Temporal Filter} To ensure the quality and realism of the paired video data, the pipeline filters out segments containing "freeze frames"—periods where one ego-vehicle is stationary causing the second scene to be static for an extended duration. This is achieved by computing frame-to-frame differences (e.g., mean squared error) and splitting or discarding segments with excessive static content. The filtering step also detects and handles timestamp gaps, further improving temporal consistency. 
The filtered video pairs are then split into fixed-length segments with configurable overlap. This segmentation ensures that each data sample is of uniform duration, facilitating batch processing and model training. Segments shorter than a minimum threshold are discarded to maintain data quality.

\mypara{Editing Instruction Generation} Finally, we use an image‑based vision‑language model (VLM)~\citep{chen2024internvl} to extract attributes for constructing the editing prompt. In Figure~\ref{App:fig-real-world-ifinder-vlm-prompt}, we show the VLM prompt used to extract global scene attributes to construct the editing prompt and video caption.

\begin{figure*}[h]
\centering
\begin{minipage}{\textwidth}
\lstset{
    basicstyle=\small\ttfamily, 
    lineskip=-2pt,     
    aboveskip=0pt,       
    belowskip=0pt,       
    breaklines=true,
    breakatwhitespace=true,
    postbreak=\mbox{\textcolor{gray}{$\hookrightarrow$}\space},
}
\begin{lstlisting}
You are an expert in autonomous driving, specializing in analyzing traffic scenes. You receive a series of traffic images from the perspective of the ego car. Your task is to describe the driving environment, focusing on weather, lighting, road layout, and environment.

It is essential that you strictly follow the rules and instructions below. Any deviation from the specified structure or format will result in an invalid output.

STRICTLY follow Rules:
 - You must strictly follow the dictionary structure provided above.
 - Only use the specified terms for weather, light, road layout, and environment. Do not create your own terms.
 - No additional information or categories should be added.
 - You should strictly follow these instructions. If an object or element is not visible or does not exist in the scene, set the value to 'None'. Ensure every field is filled with the appropriate value or 'None'.
 - STRICTLY ignore any text written on the image. 


Output the result in the following dictionary format:

{
  "surrounding_info": {
    "weather": "[e.g., 'cloudy', 'sunny', 'rainy', 'fog', 'snowy']",
    "road_layout": "[Choose from: 'straight road', 'curved road', 'intersection', 'T-junction', 'ramp']",
    "environment": "[Choose from: 'city street', 'country road', 'highway', 'residential area']",
    "sun_visibility_conditions": "[Choose from: 'clear', 'foggy', 'low visibility', 'hazy']",
    "road_condition": "[Choose from: 'wet', 'icy', 'normal', 'debris', 'potholes']",
    "surface_type": "[Choose from: 'asphalt', 'gravel', 'dirt', 'concrete']",
    "surface_color": "[Choose from: 'light grey', 'dark grey', 'black', 'brown']",
    "time_of_the_day": "[Choose from: 'morning', 'midday', 'afternoon', 'night', 'dawn', 'dusk'.]",
    "precipitation_intensity": "[Choose from: 'none', 'light', 'moderate', 'heavy', 'torrential'.]",
    "precipitation_visibility_impact": "[Choose from: 'none', 'low', 'moderate', 'high']",
    "cloud_cover": "[Choose from: 'clear', 'light', 'moderate', 'heavy'.]
    }
}
\end{lstlisting}
\end{minipage}
\caption{Annotation pipeline prompt used for the VLM.}\label{App:fig-real-world-ifinder-vlm-prompt}
\end{figure*}

\section{Synthetic Appearance Generation Details}
\label{app:synthetic_generation}

We construct synthetic appearance-editing pairs from unpaired driving videos using a two-stage generation pipeline. For each source clip, we first edit only the first frame according to a target condition, such as weather, season, or time of day. The edited frame replaces the original first frame to form a conditional input video, while the remaining frames are kept unchanged.

To construct editing prompts, we randomly sample 45-frame clips from Waymo~\citep{sun2020scalability}, Boreas~\citep{burnett_ijrr23}, and nuScenes~\citep{caesar2020nuscenes} (10k clips for each dataset). For each clip, we use InternVL~\citep{zhu2025internvl3} to identify the original condition of the first frame, including weather, time of day, and season. The target condition is randomly sampled from the remaining valid conditions after removing the original condition and impossible combinations. The candidate condition lists are:
time of day: \{\texttt{None}, dusk, night, dawn, noon, daytime\};
season: \{\texttt{None}, summer, spring, winter, autumn\};
and weather: \{\texttt{None}, sunny, snowy, rainy, foggy, cloudy\}, where \texttt{None} indicates that the corresponding attribute is not edited.

We then feed the conditional video and the source depth sequence into a fine-tuned Cosmos Transfer2.5 model with visual and depth control branches. For efficiency, we fine-tune the visual and depth branches using 480p videos with 45 frames. The edited first frame provides the target appearance condition, while depth control helps preserve scene geometry and layout. We also use negative prompts to discourage object removal, hallucination, and structural deformation.

\section{Reward Map Construction}\label{app:rewards}

To construct our fine-grained reward map $\mathbf{r} \in [-1,1]^{3 \times H \times W \times T}$, we combine complementary video-editing metrics from \citet{chen2025editboardcomprehensiveevaluationbenchmark} with custom, driving-specific semantic checks. 
We show our reward maps in Fig.~\ref{fig:reward-construction-pipeline}. These signals identify which regions provide reliable supervision for the target edit and which regions contain safety-critical traffic semantics that must be protected.
The video-editing metrics cover three dimensions: Temporal and Motion Consistency, Instruction alignment, and Subject and Scene Consistency.

\textbf{Content Preservation.}
Synthetic videos may contain unexpected scene changes, such as object removal, hallucination, or structural deformation. If directly used as supervision, these artifacts may propagate into our model. To identify such unreliable regions, we adopt the generalizable scene change detection method from~\citep{kim2025generalizablescenechangedetection}, which uses SAM features to produce scene change candidates and refines them with geometric-semantic mask matching. We generate dense masks marking regions where the synthetic target deviates structurally from the source. 

Furthermore, we design two driving aware metrics for safety critical components. 
First we target traffic lines, crosswalks, and painted arrows. We segment them with SAM3~\citep{carion2025sam3segmentconcepts} and classify each pixel into one of four states: strictly preserved, hallucinated in the target, deleted in the target, or background.
Second, we use SAM3 to extract vehicle masks in both the input and ground truth videos. 
We greedily match vehicle instances based on pixel Intersection-over-Union (IoU) and prompt a Vision-Language Model from \citep{chen2024internvl} to determine if the original body paint is accurately preserved. Together, these spatial masks heavily penalize the Content Preservation reward channel wherever source geometry is corrupted. The prompts are provided in Sec.~\ref{app:metrics}.

\textbf{Temporal Consistency.}
To measure structural fidelity across frames relative to the source video, we utilize FF-$\alpha$ and FF-$\beta$. 
FF-$\alpha$ measures intensity-based consistency using optical flow $(l_i)$ extracted from the original video to backward-warp the edited frames. Due to scenarios with significant object displacement where warping becomes unreliable (low valid pixel percentage), we also use FF-$\beta$ which compares the orientation of motion.

\textbf{Instruction Alignment.}
To assess how well our model implements the requested edits while maintaining high-quality generation, we compute a \textit{Success Rate} by measuring the CLIP cosine similarity of each edited frame $f'_i$ against both the source prompt ($p_s$) and the target prompt ($p_t$). A frame is considered successful if its alignment to $p_t$ is greater than to $p_s$. Since these are scalar metrics, their values are broadcast spatially across the frame to populate the Instruction Alignment channel of the dense reward map.

\section{Evaluation Protocol}\label{app:metrics}

To comprehensively evaluate our method, we assess editing performance across four distinct axes. We construct this protocol by combining general video-editing metrics~\citep{chen2025editboardcomprehensiveevaluationbenchmark}, advanced multimodal large language model alignment scores~\citep{ku2024viescoreexplainablemetricsconditional}, and our own driving-aware structural checks.

\begin{itemize}[leftmargin=*]
    \item \textbf{Temporal Consistency:} We quantify structural fidelity and motion stability using two complementary metrics. FF-$\alpha$ measures intensity-based consistency by backward-warping edited frames using the source video's optical flow, which is highly sensitive to minor structural changes. Since optical flow warping may be unreliable during large displacements, we also measure FF-$\beta$, which directly evaluates motion orientation consistency even when objects undergo significant positional changes.
    
    \item \textbf{Content \& ODD Preservation:} To ensure the edit does not corrupt the underlying scene, we measure preservation at both the general and driving-specific levels. For general scene preservation, we track the cross-frame stability of foreground actors (Subj.) using DINO~\citep{caron2021emergingpropertiesselfsupervisedvision} features, and background environments (Bg.) using CLIP~\citep{radford2021learningtransferablevisualmodels} features. For strict driving-aware preservation, we measure Road Line Preservation (Lane $\uparrow$) via spatial IoU of SAM3 masks. To quantify Car Appearance Preservation (Car $\uparrow$), we first use SAM3 to extract vehicle masks from both the input and generated videos. Next, we greedily match vehicle instances based on pixel Intersection-over-Union (IoU) and prompt InternVL to verify that the matched vehicles retain their original body paint. Specifically, we use the following prompt:
    \begin{quote}
    \raggedright
    \small\texttt{"Aligned crops of one vehicle: first image from the original frame, second from a generated frame. Is the body paint the same color? Ignore lighting, white balance, and window reflections. Reply with only JSON: \{"same": true\} or \{"same": false\}"}
    \end{quote}

    \item \textbf{Instruction Alignment:} We evaluate how faithfully the model executes the text prompt. The Success Rate (Succ.) calculates the frame-level frequency where the generated frame aligns better with the target prompt than the source prompt in CLIP latent space. We also report the global CLIP Similarity (CLIP) between the edited frames and the target text. Furthermore, following the evaluation framework in \citet{ku2024viescoreexplainablemetricsconditional}, we quantify complex instruction alignment using a multimodal large language model's judgment to capture nuanced text-to-video fidelity.
    
    \item \textbf{Photorealism:} We measure overall visual appeal using the LAION aesthetic predictor~\citep{schuhmann2022laion} (Aes.) and assess the absence of low-level visual distortions (e.g., blur, noise) using MUSIQ~\citep{ke2021musiq} (Img.).
\end{itemize}

\section{User Study Details}\label{app:user-study-details}

To evaluate our approach, we conducted a user study comparing \ourmethod against three baselines: CC-Edit, UniVideo, and SyntheticPip. Test samples were randomly chosen from our combined in-distribution (ID) and out-of-distribution (OOD) evaluation sets.

For each survey instance, participants were presented with a source video, a target editing prompt, and the resulting videos generated by all four methods. To prevent bias, the order of the generated videos was randomized for each question. For each trial, participants evaluated the outputs according to one criterion selected at random from the 4 criteria below. If no method was a clear winner, users were instructed to select ``Comparable.'' The criteria presented to the users were as follows:

\begin{description}[leftmargin=*, style=nextline]
    \item\textbf{Temporal Consistency}: Select the edit that has the best temporal consistency \textit{(e.g., has the least sudden jumps, flickering, appearance changes, broken motion, etc.)}.
    \item \textbf{Content Preservation}: Select the edit that best preserves the original content \textit{(e.g., preserves original car colours, road line patterns, traffic light states, building geometry, plants, etc.)}.
    \item \textbf{Instruction Following}: Select the edit that best follows the instruction \textit{(e.g., most closely matches the target editing prompt)}.
    \item \textbf{Photorealism}: Select the edit that has the best photorealism \textit{(e.g., realistic colours, lighting, natural appearance, etc.)}.
\end{description}

We polled 44 participants with each participant being provided 10 questions. Each question assessed one randomly selected criterion, yielding 429 responses in total. The overall scores were computed as the total preference pooled across criteria.

\section{Training and Inference Details}\label{app:training-inference-details}

\subsection{Training \& Architecture Details}\label{app:architecture-details}
We instantiate our framework on top of Cosmos-Transfer2.5~\cite{nvidia2026worldsimulationvideofoundation}, a diffusion transformer (DiT) that performs latent-space video denoising under multiple structured control branches. 
To achieve our goal of spatiotemporal consistency, we isolate two specific spatial controls: depth and visual (RGB). The depth branch is frozen to strictly preserve the geometric layout, using maps extracted via Depth Anything V2~\citep{depth_anything_v2}.
Although originally designed as a de-blurring control, we repurpose the visual branch by providing sharp, unblurred inputs (see ``Finetune RGB only'' in Sec.~\ref{sec:abl-main-components} for an isolated ablation of this branch).
Finally, to guide the generation, both \ourmethod and the SyntheticPip baseline are conditioned on an edited image context frame $\mathbf{c}_{\text{img}}$ (Sec.~\ref{app:image-context-training-details}), which is provided by external image editors at inference time.

 We train this model with the standard diffusion objective parameterized via $v$-prediction on our hybrid dataset across $2 \times 80$GB NVIDIA H100 GPUs with a total batch size of 2 for 28000 steps with a temporal latent length of 12 frames at 480p resolution over 48 GPU hours. For optimization, we use AdamW ($\beta_1=0.9$, $\beta_2=0.999$, weight decay of $1\text{e-}3$) with a LambdaLR scheduler, a peak learning rate of $8.63\text{e-}5$, and a 100-step warmup.

 During inference we denoise for 35 steps with reward guidance scale 2.0 and text guidance scale 2.0. We use the FlowUniPCMultistepScheduler inference scheduler. 
 During inference, we generate a total of 93 frames, comprising 85 edited video frames and 8 intermediate transition frames derived from 2 'thinking latents. We use the VAE from~\citep{BoerBohan2025TAEHV} which has a temporal compression factor of 4.

\subsection{Image Context Robustness via Augmentation}\label{app:image-context-training-details}

During training, we apply an image-context augmentation to the reference frame used for conditioning. Specifically, we set the augmentation mode to Poisson blending~\citep{perez2003poissonblending} and apply it with probability 0.75; with the remaining 0.25 probability, the selected reference frame is left unchanged. When augmentation is triggered, we sample a random temporal frame from the video to edit as the donor frame. A spatial mask is generated using a random ellipse, and the donor content is composited into the target context frame via Poisson blending. This produces a locally edited but photometrically coherent reference image that preserves global scene structure while introducing controlled local appearance variation.

The objective is to increase robustness of reference-image conditioning by exposing the model to realistic, spatially localized context perturbations rather than only clean frame-level supervision. Because the blend is constrained by an irregular elliptical mask and solved with seamless cloning, the augmented context remains visually plausible while still encouraging the model to tolerate partial mismatches between reference appearance and target trajectory.

\section{Dataset Impacts}\label{app:abl-datasets}

In Tab.~\ref{tab:data_mixture_absolute} and Tab.~\ref{tab:data_mixture_relative}, we show the impacts of our dataset construction. 
While real-world data carries groundtruth information about imaging quality and instruction alignment, due to temporal misalignments and traffic changes it cannot supervise safety critical car appearance preservation. 
Results show that training on hybrid datasets shows drastic improvements in temporal consistency and traffic semantic preservation. These gains far outweigh the minor performance drops observed in instruction alignment.

\begin{table*}[h]
\caption{\textbf{Ablation on Data Mixture.} We evaluate training exclusively on synthetic edits (Pseudo-only) versus real-world data (Real-only), compared to our hybrid approach. Results are computed on the OOD test data. Best results are \bestinline{grey bold}; runner-ups are \secondinline{light grey}.}
\centering
\footnotesize 
\setlength{\tabcolsep}{3.2pt} 
\renewcommand{\arraystretch}{0.9}
\begin{tabular}{l cc ccc cccc cc}
\toprule
& \multicolumn{2}{c}{\textbf{Temporal}} & \multicolumn{3}{c}{\textbf{Edit Align.}} & \multicolumn{4}{c}{\textbf{ODD Preservation}} & \multicolumn{2}{c}{\textbf{Quality}} \\
\cmidrule(lr){2-3} \cmidrule(lr){4-6} \cmidrule(lr){7-10} \cmidrule(lr){11-12}
\textbf{Data} & \textbf{FF-$\alpha$↓} & \textbf{FF-$\beta$↓} & \textbf{Succ.↑} & \textbf{CLIP↑} & \textbf{GPT↑} & \textbf{Subj.↑} & \textbf{Bg.↑} & \textbf{Lane↑} & \textbf{Col.↑} & \textbf{Aes.↑} & \textbf{Img.↑} \\
\midrule
Ours (Pseudo-only) & 3.3004 & 0.5402 & \best{0.9048} & \best{0.2342} & \best{7.4230} & 0.9119 & 0.9520 & 0.5615 & 0.8296 & 0.5348 & 0.5975 \\
Ours (Real-only)   & 3.4974 & \best{0.5215} & \second{0.8998} & \second{0.2309} & \second{7.3903} & \best{0.9137} & \best{0.9540} & \second{0.6246} & \second{0.8337} & \best{0.5536} & \best{0.6135} \\
\textbf{Ours} & \best{3.2185} & \second{0.5235} & 0.8762 & 0.2299 & 7.3836 & \second{0.9126} & \second{0.9525} & \best{0.6281} & \best{0.8558} & \second{0.5472} & \second{0.6030} \\
\bottomrule
\end{tabular}
\label{tab:data_mixture_absolute}
\end{table*}

\begin{table*}[h]
\caption{\textbf{Relative Improvement (\%) over Hybrid Data.} We report percentage improvement compared to our hybrid dataset for temporal stability, instruction alignment, ODD preservation, and overall video quality showing \bestinline{best} and \secondinline{runner-up} improvements. Results are computed on the OOD test data.}
\centering
\footnotesize
\setlength{\tabcolsep}{3.2pt} 
\renewcommand{\arraystretch}{0.9}
\begin{tabular}{l cc ccc cccc cc}
\toprule
& \multicolumn{2}{c}{\textbf{Temporal}} & \multicolumn{3}{c}{\textbf{Edit Align.}} & \multicolumn{4}{c}{\textbf{ODD Preservation}} & \multicolumn{2}{c}{\textbf{Quality}} \\
\cmidrule(lr){2-3} \cmidrule(lr){4-6} \cmidrule(lr){7-10} \cmidrule(lr){11-12}
\textbf{Data Variant} & \textbf{FF-$\alpha$↓} & \textbf{FF-$\beta$↓} & \textbf{Succ.↑} & \textbf{CLIP↑} & \textbf{GPT↑} & \textbf{Subj.↑} & \textbf{Bg.↑} & \textbf{Lane↑} & \textbf{Car↑} & \textbf{Aes.↑} & \textbf{Img.↑} \\
\midrule
Ours (Pseudo-only) & -2.54 & -3.20 & \best{+3.27} & \best{+1.85} & \best{+0.53} & -0.07 & -0.06 & -10.61 & -3.06 & -2.27 & -0.91 \\
Ours (Real-only)   & -8.67 & \best{+0.37} & \second{+2.69} & \second{+0.43} & \second{+0.09} & \best{+0.13} & \best{+0.16} & -0.56 & -2.59 & \best{+1.16} & \best{+1.74} \\
\textbf{Ours} & -- & -- & -- & -- & -- & -- & -- & -- & -- & -- & -- \\
\bottomrule
\end{tabular}
\label{tab:data_mixture_relative}
\end{table*}

\section{Dataset Statistics}
\label{app:dataset_statistics}

\subsection{Synthetic Paired Videos}

\begin{table}[h]
\centering
\caption{Comparison between original autonomous-driving datasets and our generated video-editing dataset. Existing datasets provide raw driving logs or perception/planning annotations, while our dataset provides paired source/target videos with explicit editing supervision.}
\label{tab:dataset_supervision_comparison}
\small
\resizebox{\linewidth}{!}{
\begin{tabular}{lccccc}
\toprule
Dataset 
& Native unit 
& Native length 
& Paired videos 
& Target diversity 
& Supervision \\
\midrule
Boreas~\cite{burnett_ijrr23} 
& Driving logs 
& Variable-length routes 
& No 
& Natural weather/season variation 
& Raw sensor data, poses \\

nuScenes~\cite{caesar2020nuscenes} 
& 1,000 scenes 
& 20s per scene 
& No 
& Natural urban conditions 
& 3D detection/tracking annotations \\

Waymo~\cite{sun2020scalability} 
& 1,150 segments 
& 20s / 200 frames per segment 
& No 
& Natural driving conditions 
& 2D/3D perception annotations \\

nuPlan~\cite{karnchanachari2024towards} 
& Driving logs 
& Large-scale variable-length logs 
& No 
& Multi-city driving scenarios 
& Planning annotations \\

\midrule
\textbf{Ours} 
& Editing pairs 
& 45 frames per clip 
& \textbf{Yes} 
& \textbf{Weather, time, season} 
& \textbf{Edited target video, reference image} \\
\bottomrule
\end{tabular}
}
\end{table}

\begin{wraptable}{R}{0.52\textwidth}
\vspace{-1.5em}
\caption{Video counts in our generated driving-video editing dataset.}
\label{tab:dataset_clip_counts}
\centering
\footnotesize
\setlength{\tabcolsep}{4pt}
\begin{tabular}{lccccc}
\toprule
Split & Boreas & nuScenes & Waymo & nuPlan & Total \\
\midrule
Train & 9,406 & 9,496 & 9,446 & -- & 28,348 \\
ID Test & 50 & 50 & 50 & -- & 150 \\
OOD Test & -- & -- & -- & 150 & 150 \\
\bottomrule
\end{tabular}
\vspace{-1em}
\end{wraptable}

\begin{table}[h]
\centering
\caption{Distribution by edited attribute for synthetic training data, in-domain test data, and OOD test data splits. Rows denote original conditions and columns denote target conditions. Clips with \texttt{None} for the corresponding attribute are excluded.}
\label{tab:attribute_distribution}
\small

{\footnotesize\textbf{Train}}

\vspace{0.2em}
\begin{minipage}{0.32\linewidth}
\centering
{\footnotesize\textbf{Weather}}
\resizebox{\linewidth}{!}{
\begin{tabular}{lrrrrr}
\toprule
Orig.$\rightarrow$Tgt. & sunny & cloudy & rainy & snowy & foggy \\
\midrule
sunny  & -- & 2708 & 2687 & 2358 & 2672 \\
cloudy & 1800 & -- & 2144 & 1688 & 2134 \\
rainy  & 714 & 801 & -- & 578 & 750 \\
snowy  & 145 & 200 & 206 & -- & 178 \\
foggy  & 171 & 182 & 187 & 149 & -- \\
\bottomrule
\end{tabular}
}
\end{minipage}
\hfill
\begin{minipage}{0.32\linewidth}
\centering
{\footnotesize\textbf{Time of Day}}
\resizebox{\linewidth}{!}{
\begin{tabular}{lrrrrr}
\toprule
Orig.$\rightarrow$Tgt. & day & noon & dawn & dusk & night \\
\midrule
day   & -- & 4794 & 4822 & 4806 & 4345 \\
dawn  & 34 & 37 & -- & 50 & 27 \\
dusk  & 426 & 426 & 427 & -- & 364 \\
night & 530 & 560 & 568 & 554 & -- \\
\bottomrule
\end{tabular}
}
\end{minipage}
\hfill
\begin{minipage}{0.32\linewidth}
\centering
{\footnotesize\textbf{Season}}
\resizebox{\linewidth}{!}{
\begin{tabular}{lrrrr}
\toprule
Orig.$\rightarrow$Tgt. & spring & summer & autumn & winter \\
\midrule
spring & -- & 312 & 435 & 471 \\
summer & 2198 & -- & 2235 & 2239 \\
autumn & 3247 & 2572 & -- & 3348 \\
winter & 1040 & 946 & 1135 & -- \\
\bottomrule
\end{tabular}
}
\end{minipage}

\vspace{0.8em}

{\footnotesize\textbf{In-Domain Test}}

\vspace{0.2em}
\begin{minipage}{0.32\linewidth}
\centering
{\footnotesize\textbf{Weather}}
\resizebox{\linewidth}{!}{
\begin{tabular}{lrrrrr}
\toprule
Orig.$\rightarrow$Tgt. & sunny & cloudy & rainy & snowy & foggy \\
\midrule
sunny  & -- & 20 & 15 & 13 & 13 \\
cloudy & 17 & -- & 9 & 8 & 8 \\
rainy  & 7 & 3 & -- & 4 & 3 \\
snowy  & 0 & 2 & 1 & -- & 0 \\
foggy  & 0 & 1 & 0 & 1 & -- \\
\bottomrule
\end{tabular}
}
\end{minipage}
\hfill
\begin{minipage}{0.32\linewidth}
\centering
{\footnotesize\textbf{Time of Day}}
\resizebox{\linewidth}{!}{
\begin{tabular}{lrrrrr}
\toprule
Orig.$\rightarrow$Tgt. & day & noon & dawn & dusk & night \\
\midrule
day   & -- & 27 & 27 & 27 & 27 \\
dusk  & 7 & 1 & 1 & -- & 0 \\
night & 8 & 0 & 0 & 0 & -- \\
\bottomrule
\end{tabular}
}
\end{minipage}
\hfill
\begin{minipage}{0.32\linewidth}
\centering
{\footnotesize\textbf{Season}}
\resizebox{\linewidth}{!}{
\begin{tabular}{lrrrr}
\toprule
Orig.$\rightarrow$Tgt. & spring & summer & autumn & winter \\
\midrule
spring & -- & 4 & 4 & 4 \\
summer & 9 & -- & 19 & 11 \\
autumn & 10 & 20 & -- & 13 \\
winter & 8 & 6 & 5 & -- \\
\bottomrule
\end{tabular}
}
\end{minipage}

\vspace{0.8em}

{\footnotesize\textbf{OOD Test}}

\vspace{0.2em}
\begin{minipage}{0.32\linewidth}
\centering
{\footnotesize\textbf{Weather}}
\resizebox{\linewidth}{!}{
\begin{tabular}{lrrrrr}
\toprule
Orig.$\rightarrow$Tgt. & sunny & cloudy & rainy & snowy & foggy \\
\midrule
sunny  & -- & 26 & 23 & 25 & 24 \\
cloudy & 24 & -- & 2 & 0 & 1 \\
\bottomrule
\end{tabular}
}
\end{minipage}
\hfill
\begin{minipage}{0.32\linewidth}
\centering
{\footnotesize\textbf{Time of Day}}
\resizebox{\linewidth}{!}{
\begin{tabular}{lrrrrr}
\toprule
Orig.$\rightarrow$Tgt. & day & noon & dawn & dusk & night \\
\midrule
day  & -- & 30 & 30 & 30 & 28 \\
dusk & 2 & 0 & 0 & -- & 1 \\
\bottomrule
\end{tabular}
}
\end{minipage}
\hfill
\begin{minipage}{0.32\linewidth}
\centering
{\footnotesize\textbf{Season}}
\resizebox{\linewidth}{!}{
\begin{tabular}{lrrrr}
\toprule
Orig.$\rightarrow$Tgt. & spring & summer & autumn & winter \\
\midrule
spring & -- & 19 & 1 & 4 \\
summer & 27 & -- & 29 & 23 \\
autumn & 2 & 9 & -- & 2 \\
\bottomrule
\end{tabular}
}
\end{minipage}

\end{table}

We construct a synthetic paired driving-video editing dataset from Boreas~\cite{burnett_ijrr23}, nuScenes~\cite{caesar2020nuscenes}, Waymo~\cite{sun2020scalability}, and nuPlan~\cite{karnchanachari2024towards}. Unlike the original autonomous-driving datasets, which mainly provide raw driving logs or perception/planning annotations, our dataset provides explicit editing supervision. Each sample contains a 45-frame source clip, an edited target clip, a natural-language instruction, and a first-frame reference image. As summarized in Table~\ref{tab:dataset_supervision_comparison}, our dataset converts existing driving logs into paired source/target videos with controllable appearance-editing targets.

Table~\ref{tab:dataset_clip_counts} reports the dataset scale. We generate 28,348 training clips from Boreas, nuScenes, and Waymo, and reserve held-out clips from the same datasets for in-domain testing. We use nuPlan only for out-of-distribution testing to evaluate cross-dataset generalization. For training, the first-frame reference image is generated with Qwen Image Editing~\cite{wu2025qwenimagetechnicalreport} for cost efficiency at scale. For both in-domain and OOD testing, the reference image is generated with Nano Banana Pro~\cite{google_nanobanana}.

The editing instructions cover global appearance changes over weather, time of day, and season. Table~\ref{tab:attribute_distribution} shows the training-set, ID and OOD testing set transition distribution for each edited attribute. Rows denote the original condition and columns denote the target condition. Clips with \texttt{None} for a given attribute are excluded from that attribute matrix, since that attribute is unchanged while another attribute is edited.

\subsection{Real-World Paired Videos}
\label{app:real_world_paired_videos}

In addition to synthetic paired videos, we also use real-world pose-paired videos from Boreas~\citep{burnett_ijrr23}. These pairs are constructed by matching repeated-route traversals according to camera pose (Sec.~\ref{app:real-world-data}).
We employ a bidirectional training strategy where the source and target roles within each pair are interchangeable. Consequently, the 33,750 unique pairs yield 67,500 directional training examples.

To report the condition-transition distribution in Table~\ref{tab:boreas_parquet_distribution} we normalize condition labels into the same taxonomy used by the synthetic dataset: \texttt{clear}$\rightarrow$\texttt{sunny}, \texttt{morning}$\rightarrow$\texttt{dawn}, \texttt{afternoon}$\rightarrow$\texttt{daytime}, and \texttt{evening}$\rightarrow$\texttt{dusk}. 

\begin{table}[h]
\centering
\caption{Condition-transition distribution of the real-world pose-paired Boreas data. Rows denote source conditions and columns denote target conditions. Counts only include clips with normalized non-\texttt{None} labels for the corresponding attribute.}
\label{tab:boreas_parquet_distribution}
\small

{\footnotesize\textbf{Real-world Paired Dataset}}

\vspace{0.2em}
\begin{minipage}{0.32\linewidth}
\centering
{\footnotesize\textbf{Weather}}
\resizebox{\linewidth}{!}{
\begin{tabular}{lrrrrr}
\toprule
Orig.$\rightarrow$Tgt. & sunny & cloudy & rainy & snowy & foggy \\
\midrule
sunny  & -- & 9641 & 1520 & 3506 & 578 \\
cloudy & 9641 & -- & 984 & 2183 & 325 \\
rainy  & 1520 & 984 & -- & 390 & 57 \\
snowy  & 3506 & 2183 & 390 & -- & 40 \\
foggy  & 578 & 325 & 57 & 40 & -- \\
\bottomrule
\end{tabular}
}
\end{minipage}
\hfill
\begin{minipage}{0.32\linewidth}
\centering
{\footnotesize\textbf{Time of Day}}
\resizebox{\linewidth}{!}{
\begin{tabular}{lrrrrr}
\toprule
Orig.$\rightarrow$Tgt. & day & noon & dawn & dusk & night \\
\midrule
day   & -- & 8116 & 4439 & 889 & 825 \\
noon  & 8116 & -- & 4216 & 806 & 800 \\
dawn  & 4439 & 4216 & -- & 558 & 550 \\
dusk  & 889 & 806 & 558 & -- & 89 \\
night & 825 & 800 & 550 & 89 & -- \\
\bottomrule
\end{tabular}
}
\end{minipage}
\hfill
\begin{minipage}{0.32\linewidth}
\centering
{\footnotesize\textbf{Season}}
\resizebox{\linewidth}{!}{
\begin{tabular}{lrrrr}
\toprule
Orig.$\rightarrow$Tgt. & spring & summer & autumn & winter \\
\midrule
spring & -- & 2933 & 3613 & 5277 \\
summer & 2933 & -- & 2500 & 4149 \\
autumn & 3613 & 2500 & -- & 4261 \\
winter & 5277 & 4149 & 4261 & -- \\
\bottomrule
\end{tabular}
}
\end{minipage}

\vspace{0.3em}
\begin{minipage}{0.95\linewidth}
\footnotesize
\emph{Note:} Some clips keep a given attribute unchanged or have missing condition annotations, so they are excluded from the corresponding transition matrix.
\end{minipage}
\end{table}

\subsection{Overall Editing Coverage}
\label{app:overall_editing_coverage}

Table~\ref{tab:edit_category_counts} summarizes how many clips edit each condition type. A single clip may edit multiple attributes simultaneously, so the weather, time-of-day, and season counts are not mutually exclusive. The synthetic splits are edited by construction, while the real-world Boreas parquet pairs may contain unchanged or missing condition attributes.

\begin{table}[t]
\centering
\caption{Number of videos editing each condition type. Counts are not mutually exclusive because one clip may edit multiple attributes.}
\label{tab:edit_category_counts}
\small
\resizebox{0.7\linewidth}{!}{
\begin{tabular}{lrrrr}
\toprule
Dataset & Weather & Time of Day & Season & Total Videos \\
\midrule
Train-Boreas & 7510 & 7529 & 7106 & 9406 \\
Train-nuScenes & 7559 & 7643 & 7139 & 9496 \\
Train-Waymo & 7590 & 7598 & 7018 & 9446 \\
ID Test-Boreas & 42 & 42 & 40 & 50 \\
ID Test-nuScenes & 42 & 41 & 40 & 50 \\
ID Test-Waymo & 42 & 42 & 40 & 50 \\
OOD-nuPlan & 125 & 121 & 120 & 150 \\
Train-Real-World-Boreas & 38448 & 42576 & 45466 & 57,966 \\
\midrule
\textbf{Total} & \textbf{61358} & \textbf{65592} & \textbf{66969} & \textbf{86614} \\
\bottomrule
\end{tabular}
}
\end{table}

Table~\ref{tab:multi_condition_counts} reports how many attributes are edited simultaneously. Most synthetic videos edit two or three attributes, providing compound appearance changes such as snowy winter dawn or rainy daytime autumn. For \texttt{Train-Real-World-Boreas}, 9,534 of 67,500 videos have no valid condition change or contain missing annotations; therefore, its total edited count is smaller than its total number of videos.

\begin{table}[t]
\centering
\caption{Number of conditions (weather, season, and time of day) edited simultaneously. \texttt{1-cat.}, \texttt{2-cat.}, and \texttt{3-cat.} indicate whether one, two, or all three attributes are changed.}
\label{tab:multi_condition_counts}
\small
\resizebox{0.6\linewidth}{!}{
\begin{tabular}{lrrrr}
\toprule
Dataset & 1-cat. & 2-cat. & 3-cat. & Total Videos \\
\midrule
Train-Boreas & 1097 & 3879 & 4430 & 9406 \\
Train-nuScenes & 1122 & 3903 & 4471 & 9496 \\
Train-Waymo & 1086 & 3960 & 4400 & 9446 \\
ID Test-Boreas & 4 & 18 & 28 & 50 \\
ID Test-nuScenes & 3 & 21 & 26 & 50 \\
ID Test-Waymo & 3 & 20 & 27 & 50 \\
OOD-nuPlan & 16 & 52 & 82 & 150 \\
Train-Real-World-Boreas & 10696 & 26016 & 21254 & 57966 \\
\midrule
\textbf{Total} & \textbf{14027} & \textbf{37869} & \textbf{34718} & \textbf{86614} \\
\bottomrule
\end{tabular}
}
\end{table}

Table~\ref{tab:target_subtype_counts} further shows the target-value distribution across all synthetic and real-world paired data. The resulting dataset covers all target weather, time-of-day, and season categories, while preserving the natural imbalance inherited from real driving logs.
\begin{table}[t]
\centering
\caption{Number of edits per target subtype across all paired data.}
\label{tab:target_subtype_counts}
\small

\begin{minipage}{0.31\linewidth}
\centering
{\footnotesize\textbf{Weather}}
\resizebox{\linewidth}{!}{
\begin{tabular}{lrrrrr}
\toprule
Target & sunny & cloudy & snowy & rainy & foggy \\
\midrule
Count & 18171 & 17115 & 10974 & 8265 & 6833 \\
\bottomrule
\end{tabular}
}
\end{minipage}
\hfill
\begin{minipage}{0.31\linewidth}
\centering
{\footnotesize\textbf{Time of Day}}
\resizebox{\linewidth}{!}{
\begin{tabular}{lrrrrr}
\toprule
Target & noon & dawn & day & dusk & night \\
\midrule
Count & 19813 & 15638 & 15276 & 7809 & 7056 \\
\bottomrule
\end{tabular}
}
\end{minipage}
\hfill
\begin{minipage}{0.31\linewidth}
\centering
{\footnotesize\textbf{Season}}
\resizebox{\linewidth}{!}{
\begin{tabular}{lrrrr}
\toprule
Target & winter & spring & autumn & summer \\
\midrule
Count & 20079 & 18656 & 14536 & 13698 \\
\bottomrule
\end{tabular}
}
\end{minipage}

\end{table}

\section{Reward Impacts}\label{app:abl-rewards}
Tab.~\ref{tab:reward_ablation_relative_ood} shows that combining the temporal consistency, instruction alignment, and content preservation reward channels during training yields the strongest overall generalization to out-of-distribution datasets. Crucially, this combined approach \textit{simultaneously} maximizes both temporal stability and content preservation, while maintaining stable imaging quality and competitive instruction alignment. The corresponding raw scores are provided in Tab.~\ref{tab:reward_ablation_raw_ood}.

\begin{table*}[h]
\caption{\textbf{Rewards: Relative Improvement over Baseline.} We evaluate the impact of individual components (content, instruction, temporal) compared to our full model and the SyntheticPip baseline. We report percentage improvement on OOD data compared to the SyntheticPip baseline. \bestinline{Best} and \secondinline{runner-up} improvements are highlighted.}
\centering
\footnotesize
\setlength{\tabcolsep}{3.2pt} 
\renewcommand{\arraystretch}{0.9}
\begin{tabular}{l cc ccc cccc cc}
\toprule
& \multicolumn{2}{c}{\textbf{Temporal}} & \multicolumn{3}{c}{\textbf{Edit Align.}} & \multicolumn{4}{c}{\textbf{ODD Preservation}} & \multicolumn{2}{c}{\textbf{Quality}} \\
\cmidrule(lr){2-3} \cmidrule(lr){4-6} \cmidrule(lr){7-10} \cmidrule(lr){11-12}
\textbf{Variant} & \textbf{FF-$\alpha$↓} & \textbf{FF-$\beta$↓} & \textbf{Succ.↑} & \textbf{CLIP↑} & \textbf{GPT↑} & \textbf{Subj.↑} & \textbf{Bg.↑} & \textbf{Lane↑} & \textbf{Car↑} & \textbf{Aes.↑} & \textbf{Img.↑} \\
\midrule
SyntheticPip (baseline) & -- & -- & -- & -- & -- & -- & -- & -- & -- & -- & -- \\
only content & \second{+5.55} & \second{+0.15} & -3.39 & \best{+0.45} & -0.20 & +1.15 & \best{+0.89} & \second{+7.70} & \second{+7.85} & -0.29 & -2.01 \\
only instruction & +4.84 & \second{+0.53} & -4.12 & \second{+0.06} & -0.77 & \second{+1.21} & \second{+0.88} & +8.14 & +7.98 & \best{+0.87} & -0.59 \\
only temporal & +5.19 & +0.13 & -3.33 & \second{+0.06} & \second{-1.41} & \best{+1.23} & +0.84 & +7.79 & \second{+8.73} & \second{+0.35} & -1.18 \\
\textbf{Ours} & \best{+7.25} & \best{+0.94} & -5.67 & -0.37 & -1.36 & +1.19 & +0.87 & \best{+8.78} & \best{+8.61} & +0.14 & \second{-1.57} \\
\bottomrule
\end{tabular}
\label{tab:reward_ablation_relative_ood}
\end{table*}

\begin{table*}[h]
\caption{\textbf{Rewards: Component Ablation.} We evaluate the impact of individual components (content, instruction, temporal) compared to our full model and the SyntheticPip baseline on OOD data. Best results are \bestinline{grey bold}; runner-ups are \secondinline{light grey}.}
\centering
\footnotesize 
\setlength{\tabcolsep}{3.2pt} 
\renewcommand{\arraystretch}{0.9}
\begin{tabular}{l cc ccc cccc cc}
\toprule
& \multicolumn{2}{c}{\textbf{Temporal}} & \multicolumn{3}{c}{\textbf{Edit Align.}} & \multicolumn{4}{c}{\textbf{ODD Preservation}} & \multicolumn{2}{c}{\textbf{Quality}} \\
\cmidrule(lr){2-3} \cmidrule(lr){4-6} \cmidrule(lr){7-10} \cmidrule(lr){11-12}
\textbf{Variant} & \textbf{FF-$\alpha$↓} & \textbf{FF-$\beta$↓} & \textbf{Succ.↑} & \textbf{CLIP↑} & \textbf{GPT↑} & \textbf{Subj.↑} & \textbf{Bg.↑} & \textbf{Lane↑} & \textbf{Col.↑} & \textbf{Aes.↑} & \textbf{Img.↑} \\
\midrule

SyntheticPip & 3.4702 & 0.5285 & \best{0.9289} & 0.2308 & \best{7.4855} & 0.9018 & 0.9443 & 0.5774 & 0.7880 & 0.5465 & \best{0.6126} \\

only content & 3.2777 & 0.5277 & 0.8973 & \second{0.2318} & \second{7.4703} & \second{0.9122} & \best{0.9527} & 0.6219 & 0.8498 & 0.5449 & 0.6003 \\

only instruction & 3.3022 & \second{0.5257} & 0.8906 & 0.2309 & 7.4279 & \best{0.9127} & \best{0.9527} & \second{0.6244} & \second{0.8509} & \best{0.5512} & \second{0.6090} \\

only temporal & 3.2901 & 0.5278 & \second{0.8980} & 0.2309 & 7.3800 & \best{0.9129} & 0.9522 & 0.6223 & \best{0.8568} & \second{0.5483} & 0.6054 \\

\textbf{Ours} & \best{3.2185} & \best{0.5235} & 0.8762 & 0.2299 & 7.3836 & 0.9126 & 0.9525 & \best{0.6281} & \best{0.8558} & 0.5472 & 0.6030 \\

\bottomrule
\end{tabular}
\label{tab:reward_ablation_raw_ood}
\end{table*}

\section{Downstream Experiment Details}
\label{app:downstream-details}

The SeRM dataset only provides annotations for cropped bottom-road images, but we aim to study segmentation performance on full-resolution images. We first project the annotations onto full-resolution images (Section~\ref{app:serm_processing}), then generate edited video clips (Section~\ref{app:serm_clip_construction}), and finally perform training (Section~\ref{app:serm_training}).

\subsection{SeRM Full-Resolution Pair Reconstruction}
\label{app:serm_processing}

The original SeRM release provides two disjoint image sources: cropped training images with pixel-wise lane-marking annotations, and full-resolution mapping frames recorded across three routes: Sangam, Ilsan, and Sangam\_Large. The training images have resolution $320\times1280$, while the mapping frames have resolution $672\times1280$. Each training crop corresponds to the bottom region of a full mapping frame, starting at row $y=352$. We reconstruct full-resolution image--annotation pairs by recovering this correspondence and padding the cropped labels back to the full image canvas.

We first obtain an approximate train-to-full-frame correspondence using compact image features. For every training image and candidate full frame, we crop the bottom $320$ rows when needed, downsample the image to $96\times24$ grayscale, and flatten it into a 2,304-dimensional feature vector. We compute an average perceptual hash for each image and retrieve the top candidate full frames by Hamming distance. These candidates are then re-ranked by the mean-squared error of z-score-normalized features, yielding an initial mapping from each training crop to a full-resolution frame.

We then verify the mapping using exact pixel-level hashing. All full mapping frames are concatenated into a single ordered sequence following the route order Sangam $\rightarrow$ Ilsan $\rightarrow$ Sangam\_Large. For each full frame, we decode the image into raw RGB bytes and crop the bottom region using the fixed crop window $(x,y,w,h)=(0,352,1280,320)$. We compute an MD5 hash for this decoded crop and compare it with the MD5 hash of each decoded training image. Matching is performed with a forward temporal cursor, so accepted matches must respect the original recording order. This produces a verified mapping where every retained training crop has an exact pixel match in the full-frame sequence.

Finally, we export the feasible subset as full-resolution image--annotation pairs. For each verified match, we copy the $672\times1280$ full RGB frame as the image. We construct the corresponding $672\times1280$ semantic label map by placing the original $320\times1280$ annotation into rows $352$--$671$ of a zero-initialized canvas. The top $352$ rows are assigned label $0$, indicating unlabeled/background regions. We also save the verified mapping file to keep the route, frame index, and filename provenance for each reconstructed pair. 

After this preprocess, we get about 100 clips with annotations.

\subsection{Video Clip Construction and Target Assignment}
\label{app:serm_clip_construction}

Starting from the verified full-resolution SeRM pairs, we partition the ordered sequence into non-overlapping 93-frame clips at 10 fps, corresponding to approximately 9.3 seconds per clip. A trailing partial clip is retained only if it contains at least 46 frames; otherwise it is discarded.

We assign each clip a target appearance condition sampled from weather, season, and time-of-day attributes. The prepared clips are then used for video generation for both SyntheticPip and our model. For SyntheticPip, we generate videos following Sec.~\ref{sec:data-pseudo-gt}. 

\begin{figure}[ht]
    \centering
    \includegraphics[width=\textwidth]{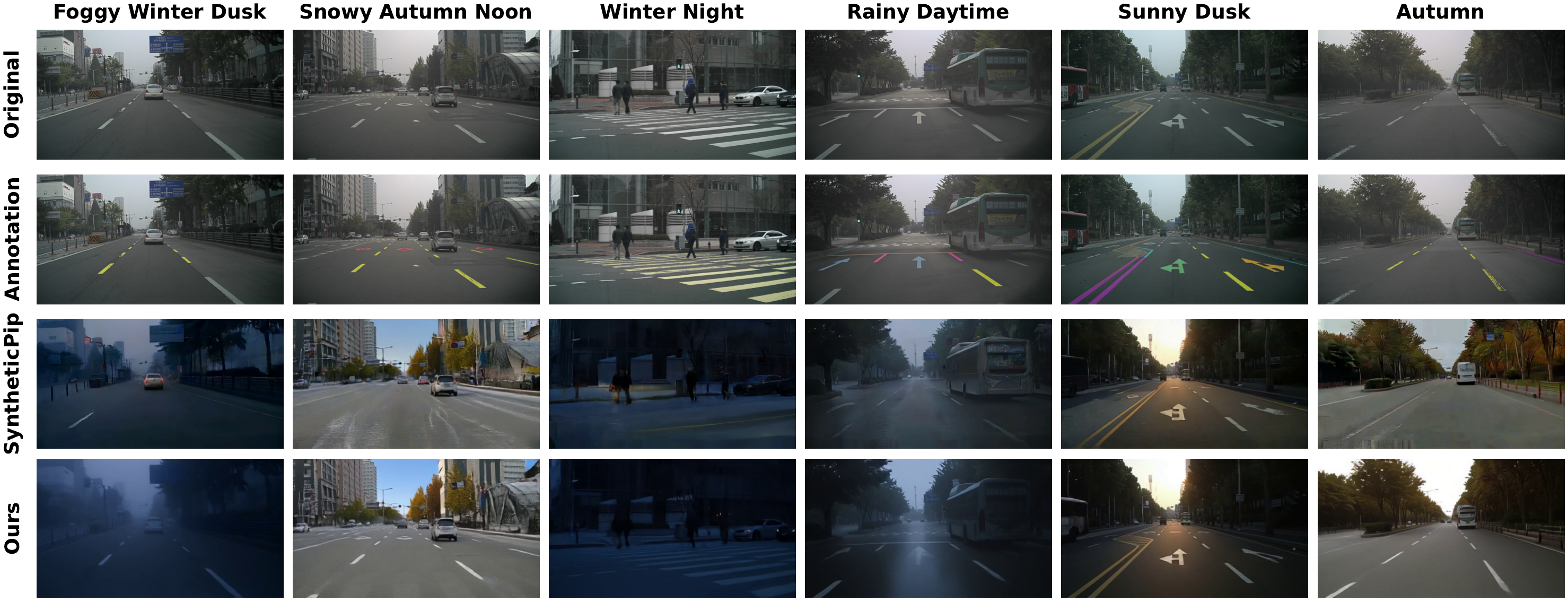}
    \caption{Qualitative comparison on SeRM lane marking segmentation. Each column shows one target scene, with rows corresponding to the original image, ground-truth annotation overlay, SyntheticPip-augmented training data, and our augmented data. 
    }
    \label{fig:downstream-task}
\end{figure}

Fig.~\ref{fig:downstream-task} visualizes the generated augmentation data used for downstream training, where our method better preserves lane markings while SyntheticPip often distorts or removes them.

Exclude 8 thinking frames for each clip, we get $100*85=8500$ frames. Because global editing should not change the scene, the generated frames can share the same annotation with the original frames.

\subsection{Training}
\label{app:serm_training}

We use PIDNet~\citep{xu2023pidnet} pretrained on CityScapes~\citep{cordts2015cityscapes} as our initial model. Then we finetuned it under following 3 settings:
(1) only original preprocessed 8500 frames;
(2) preprocessed 8500 frames and generated 8500 frames from SyntheticPip with original annotation;
(3) preprocessed 8500 frames and generated 8500 frames from our method with original annotation.
Low quality generation results which remove or distort lane markings may misguide the model, as results shown in Fig.~\ref{fig:downstream-task}.

\section{Offline Reinforcement Learning}
\label{app:offline_RL_expanded}

While neither real-world nor synthetic pairs are perfect, we treat them jointly as off-policy rollouts of an imperfect behavior policy $\mu_{\text{ref}}$ defined by our hybrid dataset $\mathcal{D}$, and
drawing on the offline RL perspective of CFGRL, we use reward-conditioned fine-tuning to learn an editing policy $\mu_\theta$ from this data.
We define $\mathbf{o}$ as the conditioning inputs (i.e., the source video and text prompt) and the action $\mathbf{a} \equiv \mathbf{x}^t$ as the generated edited video clip.

\mypara{CFGRL offline RL} We build on CFGRL~\cite{frans2025diffusion}, an offline RL framework that performs policy improvement via advantage conditioning. 
Under the standard regularized RL objective,
\begin{equation}
    J(\mu_\theta) = \mathbb{E}_{\tau \sim p(\tau \mid \mu_\theta)}\Big[\sum_t \gamma^t r(\mathbf{o}_t, a_t)\Big] - \beta \mathbb{E}_{\mathbf{o}\sim p_\mu(\mathbf{o})}[D_{\text{KL}}(\mu_\theta (\mathbf{a} \mid \mathbf{o}) \,\|\,  \mu_{\text{ref}} (\mathbf{a} \mid \mathbf{o}) )]
\end{equation}
the optimal improved policy can be viewed as a probabilistic (multiplicative) adjustment to the reference and takes the form 
$\hat{\mu}(\mathbf{a}\mid\mathbf{o}) \propto \mu_{\text{ref}}(\mathbf{a}\mid\mathbf{o}) \exp(A^{\mu_{\text{ref}}}(\mathbf{o},\mathbf{a})/\beta)$,
where $A^{\mu_{\text{ref}}}$ is the advantage of action $\mathbf{a}$ under $\mu_{\text{ref}}$, and $\beta$ is a hyperparameter \cite{frans2025diffusion}. 
Following~\cite{frans2025diffusion, intelligence2025pi06vlalearnsexperience}, the product policy can similarly be defined as $\hat{\mu}(\mathbf{a}\mid\mathbf{o}) \propto \mu_{\text{ref}}(\mathbf{a}\mid\mathbf{o}) p(I \mid A^{\mu_{\text{ref}}}(\mathbf{o},\mathbf{a}))^{\beta}$
where $p(I \mid A^{\mu_{\text{ref}}}(\mathbf{o},\mathbf{a})) = f(A^{\mu_{\text{ref}}}(\mathbf{o},\mathbf{a})) / \int f(A^{\mu_{\text{ref}}}(\mathbf{o},\mathbf{a}')) d\mathbf{a}'$ is the probability of any action improving over $\mu_\text{ref}$ for a non-negative monotonically increasing function $f$ of the advantage. Under this formulation, the idealized policy, $\hat{\mu}$ results in an improvement over $\mu_{\text{ref}}$.

Bayes rule gives $p(I \mid A^{\mu_{\text{ref}}}(\mathbf{o}, \mathbf{a})) \propto \frac{\mu_{\text{ref}}(\mathbf{a}\mid I, \mathbf{o})}{\mu_{\text{ref}}(\mathbf{a}\mid\mathbf{o})} $ leading to an alternative form of the improved policy
\begin{equation}
\hat{\mu}(\mathbf{a}\mid\mathbf{o}) \propto \mu_{\text{ref}}(\mathbf{a}\mid\mathbf{o}) \left( \frac{\mu_{\text{ref}}(\mathbf{a}\mid I, \mathbf{o})}{\mu_{\text{ref}}(\mathbf{a}\mid\mathbf{o})} \right)^{\beta},
\label{eq:cfgrl-policy-app}
\end{equation}
This can be written in terms of an improvement indicator $I = \mathbb{1}[A^{\mu_{\text{ref}}} > \epsilon]$, which avoids the need to explicitly model the advantage.
The improved policy $\hat{\mu}$ is approximated by training $\mu_\theta$ to represent both the unconditional and indicator-conditional distributions.
We apply insights from classifier-free guidance and randomly dropping $I$ during training to learn the unconditional and indicator-conditional distributions
\begin{equation}
\min_\theta \; \mathbb{E}_{\mathcal{D}}\Big[ -\log \mu_\theta(\mathbf{x}^t \mid \mathbf{o}) - \alpha \log \mu_\theta(\mathbf{x}^t \mid I, \mathbf{o}) \Big].
\label{eq:cfgrl-loss-app}
\end{equation}
where $\alpha$ is a trade-off hyperparameter following~\cite{intelligence2025pi06vlalearnsexperience}. 
At inference, we sample with $I = 1$ to favor outputs associated with higher reward.

\section{Reward Granularity}
\label{app:dense_scalar_reward}

To isolate the effect of dense reward conditioning, we train a scalar-reward baseline by aggregating the temporal-consistency, content-preservation, and instruction-alignment signals used by \ourmethod. We compare this baseline with our dense reward conditioning on the in-distribution evaluation set. As shown in Tab.~\ref{tab:dense_scalar_reward}, dense reward conditioning improves structural fidelity, with relative gains of 18.81\% in lane preservation and 3.51\% in car preservation. It also improves temporal consistency measured by FF-$\beta$ and visual quality measured by Aes.\ and Img., while the remaining metrics exhibit small differences. 

\begin{table}[h]
\centering
\caption{Dense versus scalar reward conditioning on the in-distribution evaluation set. Relative changes are computed with respect to the scalar-reward baseline; negative changes indicate improvements for FF-$\alpha$ and FF-$\beta$.}
\label{tab:dense_scalar_reward}
\resizebox{\textwidth}{!}{
\begin{tabular}{lcccccccccc}
\toprule
Method & FF-$\alpha$ $\downarrow$ & FF-$\beta$ $\downarrow$
& Succ.\ $\uparrow$ & CLIP $\uparrow$ & Lane $\uparrow$
& Car $\uparrow$ & Subj.\ $\uparrow$ & Bg.\ $\uparrow$
& Aes.\ $\uparrow$ & Img.\ $\uparrow$ \\
\midrule
Scalar reward
& \textbf{3.3313} & 0.5619 & \textbf{0.9205}
& \textbf{0.2494} & 0.4078 & 0.8068 & 0.9141
& \textbf{0.9528} & 0.4795 & 0.5785 \\
Dense reward (ours)
& 3.3539 & \textbf{0.5428} & 0.9155
& 0.2492 & \textbf{0.4845} & \textbf{0.8351}
& \textbf{0.9156} & 0.9520 & \textbf{0.4928} & \textbf{0.5873} \\
\midrule
Relative change
& +0.68\% & -3.40\% & -0.54\% & -0.08\%
& +18.81\% & +3.51\% & +0.16\% & -0.08\%
& +2.77\% & +1.52\% \\
\bottomrule
\end{tabular}
}
\end{table}

\newpage

\section*{NeurIPS Paper Checklist}

\begin{enumerate}

\item {\bf Claims}
    \item[] Question: Do the main claims made in the abstract and introduction accurately reflect the paper's contributions and scope?
    \item[] Answer: \answerYes{} % Replace by \answerYes{}, \answerNo{}, or \answerNA{}.
    \item[] Justification: The abstract and introduction sections explicitly state the main contribution (Comprehensive driving dataset, reward guidance during training, temporal augmentation frames) and the results sections back up each of these claims with empirical evidence.

\item {\bf Limitations}
    \item[] Question: Does the paper discuss the limitations of the work performed by the authors?
    \item[] Answer: \answerYes{} % Replace by \answerYes{}, \answerNo{}, or \answerNA{}.
    \item[] Justification: We add a limitations section in the appendix.

\item {\bf Theory assumptions and proofs}
    \item[] Question: For each theoretical result, does the paper provide the full set of assumptions and a complete (and correct) proof?
    \item[] Answer: \answerNA{} % Replace by \answerYes{}, \answerNo{}, or \answerNA{}.
    \item[] Justification: The paper is empirical; it introduces no theorems nor formal proofs.

    \item {\bf Experimental result reproducibility}
    \item[] Question: Does the paper fully disclose all the information needed to reproduce the main experimental results of the paper to the extent that it affects the main claims and/or conclusions of the paper (regardless of whether the code and data are provided or not)?
    \item[] Answer: \answerYes{} % Replace by \answerYes{}, \answerNo{}, or \answerNA{}.
    \item[] Justification: We provide our training setting details in the appendix. We also provide details on our dataset construction and reward map construction in the methods and appendix sections.

\item {\bf Open access to data and code}
    \item[] Question: Does the paper provide open access to the data and code, with sufficient instructions to faithfully reproduce the main experimental results, as described in supplemental material?
    \item[] Answer: \answerNo{} % Replace by \answerYes{}, \answerNo{}, or \answerNA{}.
    \item[] Justification: Due to company restrictions.

\item {\bf Experimental setting/details}
    \item[] Question: Does the paper specify all the training and test details (e.g., data splits, hyperparameters, how they were chosen, type of optimizer) necessary to understand the results?
    \item[] Answer: \answerYes{} % Replace by \answerYes{}, \answerNo{}, or \answerNA{}.
    \item[] Justification: We provide the data statistics, network architecture, and all training and inference hyperparameters in section 2, section 3 and the appendix.

\item {\bf Experiment statistical significance}
    \item[] Question: Does the paper report error bars suitably and correctly defined or other appropriate information about the statistical significance of the experiments?
    \item[] Answer: \answerYes{} % Replace by \answerYes{}, \answerNo{}, or \answerNA{}.
    \item[] Justification: We report error bars on our results in Table~\ref{tab:main-results-raw} . In addition, we evaluate our method across 2 datasets with 150 examples each and repeated over multiple ablation experiments. The consistent improvements observed across these diverse settings suggest that the reported gains are unlikely to be due to random variation.

\item {\bf Experiments compute resources}
    \item[] Question: For each experiment, does the paper provide sufficient information on the computer resources (type of compute workers, memory, time of execution) needed to reproduce the experiments?
    \item[] Answer: \answerYes{} % Replace by \answerYes{}, \answerNo{}, or \answerNA{}.
    \item[] Justification: In the Appendix we give our training GPU hours.
    
\item {\bf Code of ethics}
    \item[] Question: Does the research conducted in the paper conform, in every respect, with the NeurIPS Code of Ethics \url{https://neurips.cc/public/EthicsGuidelines}?
    \item[] Answer: \answerYes{} % Replace by \answerYes{}, \answerNo{}, or \answerNA{}.
    \item[] Justification: The work uses only synthetic simulation data and public driving scene images involving no personal or sensitive information, so it aligns with the NeurIPS Code of Ethics.

\item {\bf Broader impacts}
    \item[] Question: Does the paper discuss both potential positive societal impacts and negative societal impacts of the work performed?
    \item[] Answer: \answerYes{} % Replace by \answerYes{}, \answerNo{}, or \answerNA{}.
    \item[] Justification: This is discussed in the Appendix.
    
\item {\bf Safeguards}
    \item[] Question: Does the paper describe safeguards that have been put in place for responsible release of data or models that have a high risk for misuse (e.g., pre-trained language models, image generators, or scraped datasets)?
    \item[] Answer: \answerNA{} % Replace by \answerYes{}, \answerNo{}, or \answerNA{}.
    \item[] Justification: Our paper poses no such risks.

\item {\bf Licenses for existing assets}
    \item[] Question: Are the creators or original owners of assets (e.g., code, data, models), used in the paper, properly credited and are the license and terms of use explicitly mentioned and properly respected?
    \item[] Answer: \answerYes{} % Replace by \answerYes{}, \answerNo{}, or \answerNA{}.
    \item[] Justification: All employed code and datasets used in this work are properly cited, and their licenses permit use for research purposes.

\item {\bf New assets}
    \item[] Question: Are new assets introduced in the paper well documented and is the documentation provided alongside the assets?
    \item[] Answer: \answerNA{} % Replace by \answerYes{}, \answerNo{}, or \answerNA{}.
    \item[] Justification: We do not release new assets.

\item {\bf Crowdsourcing and research with human subjects}
    \item[] Question: For crowdsourcing experiments and research with human subjects, does the paper include the full text of instructions given to participants and screenshots, if applicable, as well as details about compensation (if any)? 
    \item[] Answer: \answerYes{} % Replace by \answerYes{}, \answerNo{}, or \answerNA{}.
    \item[] Justification: We provide the user study details and all prompts in the appendix. The videos were randomly sampled from the evaluation set and results discussed in Section 4 and the Appendix. No compensation was involved.

\item {\bf Institutional review board (IRB) approvals or equivalent for research with human subjects}
    \item[] Question: Does the paper describe potential risks incurred by study participants, whether such risks were disclosed to the subjects, and whether Institutional Review Board (IRB) approvals (or an equivalent approval/review based on the requirements of your country or institution) were obtained?
    \item[] Answer: \answerNo{} % Replace by \answerYes{}, \answerNo{}, or \answerNA{}.
    \item[] Justification: While we conducted a crowdsourced user study to evaluate visual preferences, we did not seek formal IRB approval. The study consisted solely of a standard A/B visual comparison task of generated driving videos. It posed minimal to no risk to participants, did not involve deception, and collected no Personally Identifiable Information (PII) or demographic data. Participants were fully informed of the task's nature prior to accepting it

\item {\bf Declaration of LLM usage}
    \item[] Question: Does the paper describe the usage of LLMs if it is an important, original, or non-standard component of the core methods in this research? Note that if the LLM is used only for writing, editing, or formatting purposes and does \emph{not} impact the core methodology, scientific rigor, or originality of the research, declaration is not required.
    %this research? 
    \item[] Answer: \answerNA{} % Replace by \answerYes{}, \answerNo{}, or \answerNA{}.
    \item[] Justification:  The core method development in this research does not involve LLMs as any important, original, or non-standard components.

\end{enumerate}

\end{document}